\PassOptionsToPackage{table}{xcolor}
\documentclass[subscriptcorrection,upint,varvw,barcolor=Goldenrod3,mathalfa=cal=euler,balance,hyphenate,french,nolists,nocopyright]{asmejour} %

\usepackage{multirow}
\usepackage{booktabs}
\usepackage{adjustbox}
\usepackage{balance}
\usepackage{xurl}
\newcommand{\coderepo}{\url{https://github.com/Chani-song/WAFER}}

\hypersetup{%
	pdfauthor={Chan-Eui Song, Jeremy J. Michalek, Christopher McComb},
    pdftitle={LLM-Assisted Optimization for Simulation-Based Redesign},
	pdfkeywords={large language models, design optimization, engineering redesign, simulation-based design, black-box optimization},
	pdfsubject = {Journal of Mechanical Design},
}

\JourName{Mechanical Design}%

\begin{document}

\SetAuthorBlock{Chan-Eui Song}{Department of Mechanical Engineering\\
Carnegie Mellon University\\ Pittsburgh, Pennsylvania, USA\\ email: chaneuis@andrew.cmu.edu}

   \SetAuthorBlock{Jeremy J. Michalek}{
Fellow ASME\\ Department of Mechanical Engineering\\ Department of Engineering and Public Policy\\ Dept. of Civil \& Environmental Engineering\\ Carnegie Mellon University\\ Pittsburgh, Pennsylvania, USA\\ email: jmichalek@cmu.edu}

\SetAuthorBlock{Christopher McComb\CorrespondingAuthor}{
Member ASME\\ Department of Mechanical Engineering\\ Carnegie Mellon University\\ Pittsburgh, Pennsylvania, USA\\ Email: ccm@cmu.edu }

\title{LLM-Assisted Optimization for Simulation-Based Redesign}

\keywords{Artificial Intelligence, Design Optimization, Energy Systems Design, Machine Learning, Simulation-Based Design}

\begin{abstract}

Engineering design frequently involves redesigning a prior solution to adapt to changes in context, such as material prices or system requirements. Redesign often leverages simulators too complex and/or computationally intensive for traditional numerical solvers, which lack engineering knowledge and context, to ensure global solution quality. Large language models (LLMs) can interpret semantic information, including engineering knowledge about the system and changes in context, to help guide search, and recent studies have shown that incorporating this knowledge into optimization can improve search performance. However, how best to combine LLMs with numerical solvers in redesign remains unclear. We develop and test an integrated LLM-optimization approach for redesign and compare three methods for leveraging LLMs (warm start, variable fixing and ReAct) and four black-box solvers (Bayesian optimization, genetic algorithms, mesh adaptive direct search and random search) on an electric vehicle battery pack cost minimization case study across four types of redesign scenarios. We find that the integrated approach can match feasible design discovery of the numerical-solver baselines with 80\% fewer simulator evaluations while tripling the fraction of runs that find near-optimal solutions. We find that the best LLM-solver integration approach and the way the prior solution is leveraged depend on the structure of the design requirement change, and that the benefit of increasing LLM model size grows with the extent to which the LLM controls the search. These results show clear potential for LLMs in redesign tasks for guiding or replacing numerical solvers.
\end{abstract}

\date{}%

\maketitle %

\section{Introduction}
\label{sec_introduction}

In practical engineering design, developing a satisfactory product often requires repeated cycles of evaluation and modification \cite{pahl2007engineeringdesign}. As this process continues, changes in requirements, subsystem interfaces, material availability, or system integration can make an existing design unsuitable for the updated design context. Adapting an existing design to such changes is commonly treated as redesign, and this redesign effort can consume substantial engineering capacity, development time, and cost \cite{terwiesch1999engineering,clarkson2004change,burggraf2024engineeringchange}. In the automotive industry, for example, more than 1,000 engineering change orders per month and process costs of approximately 20,000--50,000 Euros per change have been reported at a major manufacturer \cite{wasmer2011engineeringchange}. Individual engineering changes can also require several weeks or months to resolve \cite{terwiesch1999engineering}. Reducing the effort required for redesign can therefore shorten an important and costly part of the overall engineering design process.

One way to reduce this redesign effort is to use computational methods to automate the evaluation and improvement of candidate designs \cite{song2022drone,regenwetter2022biked,song2025mechanism,chen2021padgan}. In many practical optimization tasks, however, the relationship between design variables and design performance cannot be written in a compact analytical form that is convenient for direct optimization. Engineers therefore evaluate candidate designs through simulation, where physical models, numerical iterations, empirical correlations, calibrated parameters, and lookup structures may all contribute to the predicted performance \cite{talgorn2015surrogate,larson2019dfo}. When simulation is placed inside an optimization loop, the resulting process is commonly described as simulation-based design optimization. Because the internal relationships of the simulator can be difficult or impractical to expose directly to the optimizer, these problems are often treated as black-box optimization problems even when much of the underlying engineering physics is known \cite{talgorn2015surrogate,rios2013dfo}. When the design context later changes while a previous design remains available, the same setting becomes a reoptimization problem in which information from the previous design may still be useful.

Solving this black-box reoptimization problem can require many additional simulator evaluations. Numerical methods such as Bayesian optimization, evolutionary algorithms, and direct search address this problem by using information collected from previous evaluations to guide subsequent search \cite{jones1998ego,larson2019dfo,rios2013dfo}. Their search decisions are therefore driven primarily by numerical observations collected during optimization, and previous evaluations can become obsolete and/or misleading depending on the changes in design context for redesign. But the redesign context provides another source of information before many of these evaluations are performed. Engineers often know what changed in the design context, which physical relationships are relevant to that change, and which parts of the previous design are likely to remain useful. Using this semantic engineering information during reoptimization could therefore provide information that is not captured in the numerical evaluation history alone.

Large language models (LLMs) provide a possible way to incorporate semantic engineering information into optimization. LLM agents can interpret textual problem descriptions, generate structured actions, and interact with external computational tools \cite{yao2023react,schick2023toolformer}. Recent studies have explored these capabilities in optimization: LLMs have been used for iterative solution generation, warm starting, and context-informed candidate selection to improve numerical optimization \cite{yang2024opro,liu2024llambo}. Hybrid optimization methods have also combined LLM-derived prior knowledge with Bayesian optimization, motivating the separation of semantic prior reasoning from numerical search \cite{cisse2026closedloop,gupta2025llmbo}.

Similar capabilities have been explored in engineering design: LLM agents have been used to coordinate engineering analysis and simulation tools and to improve designs through repeated simulator feedback \cite{pradasgomez2026ductile,xu2026designbench}. Recent studies have further examined the effects of model scale and agent architecture in LLM-based engineering design workflows \cite{jadhav2026mechanicaldesigner,massoudi2026agentic,zhu2026mecsoagent,xu2026crdal,song2026agents4experiments}. Together, these studies show that LLMs can contribute at multiple stages of optimization and engineering design. Song et al.~\cite{song2026dynamic} showed that feedback-driven LLM search and LLM-generated warm starts can exhibit different strengths relative to conventional numerical optimization when the design context changes. However, it remains unclear how semantic engineering information about an existing design and a changed design context should be used during simulator-driven reoptimization.

This question can be studied by placing semantic guidance at different stages of the reoptimization process. Existing reoptimization methods commonly reuse information from a previous solution through mechanisms such as warm starting or variable fixing \cite{gamrath2015reoptimization,patel2024reoptimization,helber2010fixoptimize,poloczek2016warmstarting}. Recently, Ye et al.~\cite{ye2026reoptimization} introduced LLM-guided reoptimization for explicit mathematical optimization models, where the LLM translates changed requirements into structured model updates and selects the reoptimization strategy that the solver uses. In simulator-driven engineering design, however, the governing engineering relationships may instead remain embedded inside an external simulator and unavailable for direct modification. Semantic information can still influence the search by guiding the initial designs, identifying parts of the reference design that should be kept, or interpreting simulator feedback during iterative search. These possibilities motivate the three LLM roles considered in this study (warm starting, variable fixing, and ReAct search). We study these roles within a framework that we call Workflows for AI-Facilitated Engineering Redesign (WAFER).

To compare these roles under different redesign conditions, we consider a case study of lithium ion battery pack design for electric vehicles. Battery designs can require revision when material prices and supply conditions change \cite{cheng2024supplychain}, when the physical space available to the pack changes \cite{widyantara2022batterypacking}, or when vehicle level requirements are revised, such as when a pack designed for one vehicle model is redesigned for use in another model, model year, or trim level. Fast charging introduces another redesign condition because charging performance depends on electrochemical transport and polarization within the cell \cite{weiss2021fastcharging,lei2023fastcharging}. These changes require different types of engineering information and therefore provide a useful setting for examining whether the value of semantic guidance depends on the redesign scenario. We evaluate the resulting design problems using the Battery Performance and Cost (BatPaC) model developed by Argonne National Laboratory \cite{knehr2022batpac} and use the model as an external simulator throughout the study.

\begin{figure*}
    \centering
    \includegraphics[width=\linewidth]{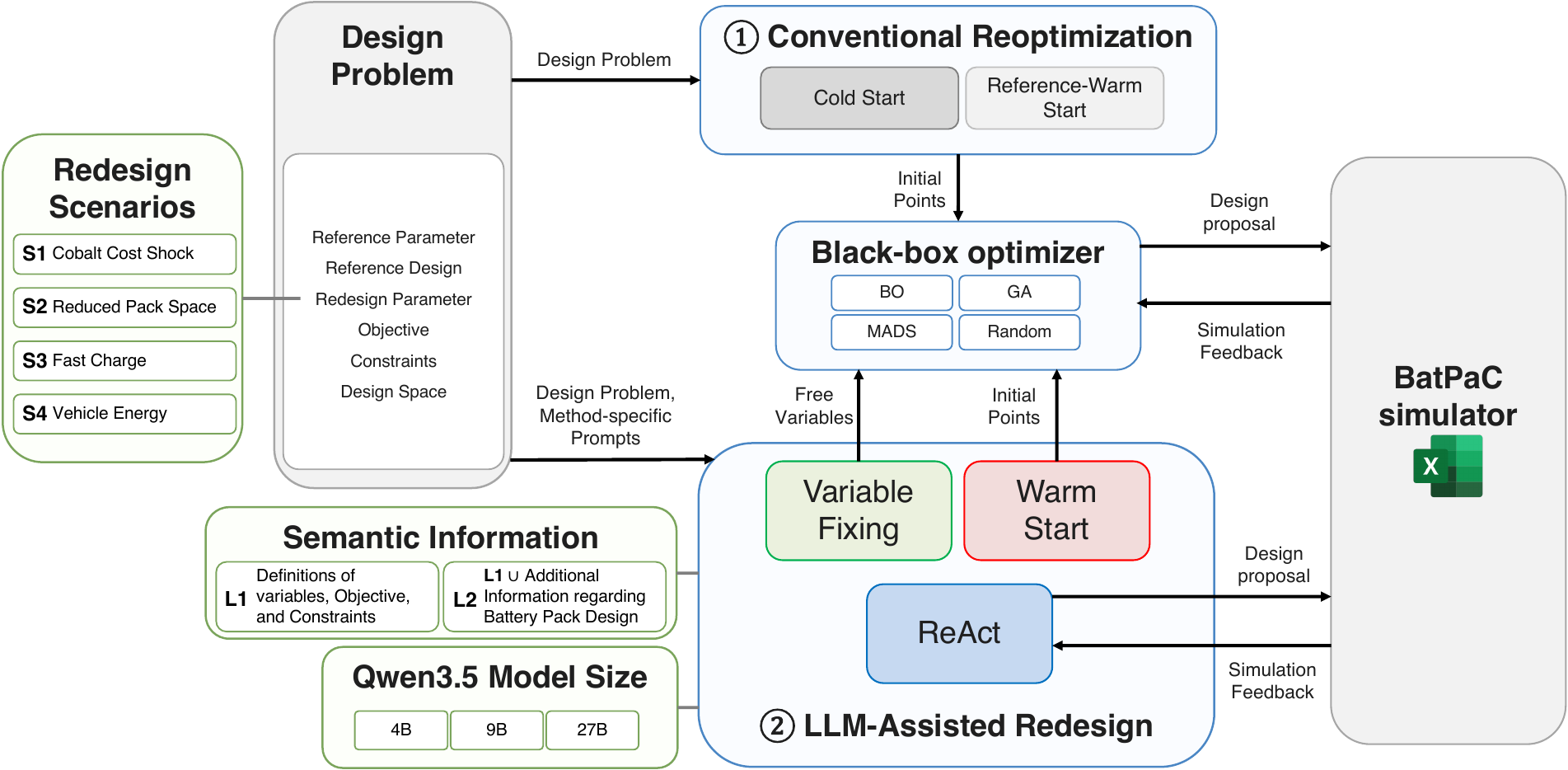}
    \caption{Overview of the redesign experiments. Each run combines one redesign scenario (S1 to S4), one semantic-information level (L1 or L2), one model size (4B, 9B, or 27B), one redesign method, and one numerical optimizer where applicable (Bayesian optimization, genetic algorithm, MADS, or random search). The three panels outlined in green contain these experimental factors. The gray panels contain the design problem and the BatPaC simulator, which every run shares. The arrows show the information passed between components. Group 1 contains the two conventional baselines (the cold start and the reference-warm start), and group 2 contains the three WAFER methods. The three method boxes are filled with the colors used for these methods in all later figures: green for variable fixing, red for warm starting, and blue for ReAct.}
    \label{fig_framework}
\end{figure*}

Using this common testbed, we study how the placement of LLM guidance affects engineering redesign performance, and Fig.~\ref{fig_framework} summarizes the framework. Starting from a common reference design, we construct four redesign scenarios involving changes to material cost, available pack space, fast charging, and vehicle energy requirements. We then compare warm starting, variable fixing, and ReAct with numerical optimization under a common simulator evaluation budget, using three language model sizes. Based on this framework, we address three research questions. 
\begin{enumerate}
    \item To what extent does LLM-assisted reoptimization improve feasible-design discovery and final design quality relative to conventional black-box optimization under a fixed simulator-evaluation budget? 
    \item What characteristics of a redesign scenario determine which LLM-assisted methods are most effective, and what mechanisms explain differences across scenarios? 
    \item How does the benefit of increased LLM size relate to the degree of control it exerts on search, and how does that benefit compare to cost in inference time and token use?
\end{enumerate}

This work makes three main contributions to LLM-assisted engineering redesign. First, we show that LLM-assisted reoptimization reaches a feasible redesign in far fewer simulator evaluations and ends at a lower cost than cold-start and reference-warm numerical optimization under a common simulation budget. Second, we show that the most effective LLM-assisted method differs across redesign scenarios, and we explain the difference through the stage of the optimization at which each strategy uses the language model. Variable fixing changes the dimension of the search, warm starting directs the search toward promising regions, and ReAct can change the search strategy at every step. Third, we show that the benefit of a larger model grows with the extent to which the model controls the search. It is largest for ReAct and inconsistent for the two strategies that use a numerical optimizer for the remaining search. We also report the inference time and token use that each strategy and model size requires, so that these gains can be weighed against their computational cost. Together, these results establish LLM-assisted reoptimization as a practical approach for reducing the cost of simulator-driven engineering redesign and provide guidance for selecting how semantic reasoning should be integrated into the redesign process.

\section{Battery-Pack Redesign Problem}
\label{sec:problem}

In this section, we define the electric vehicle (EV) battery-pack design problem used as the redesign testbed in this study. Battery-pack design couples decisions across the cell, module, and pack levels through electrochemical, electrical, and packaging relationships, while the design conditions for the pack can change as the vehicle design progresses. We formulate a common reference optimization problem, obtain the reference design used throughout the study, and construct four redesign scenarios that change distinct aspects of the reference condition.

\subsection{Battery-Pack Design Problem}
\label{sec:battery_testbed}

We build upon the battery-pack cost-minimization problem of Song et al.~\cite{song2026dynamic}, which uses BatPaC v5.0 \cite{knehr2022batpac} as a simulator. BatPaC evaluates coupled cell-, module-, and pack-level design decisions and returns battery cost and performance quantities, and all methods access it through the same external input-output interface.

\begin{figure*}[t]
    \centering
    \includegraphics[width=\linewidth]{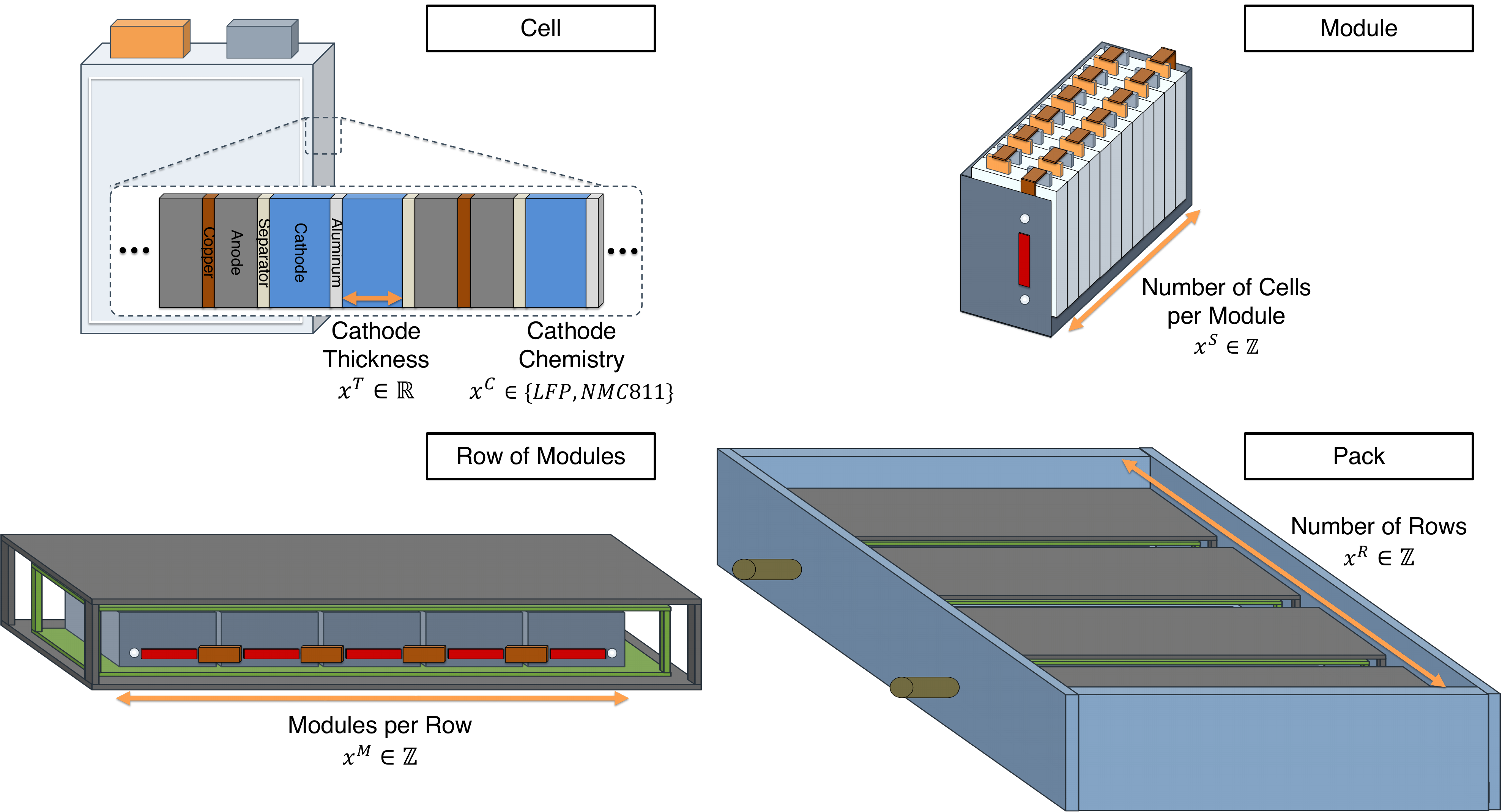}
    \caption{Battery-pack design hierarchy in BatPaC and the optimization variables. The cathode chemistry $x^{\mathrm{C}}$ and the positive-electrode thickness $x^{\mathrm{T}}$ are defined at the cell level, $x^{\mathrm{S}}$ gives the number of cells per module, $x^{\mathrm{M}}$ the number of modules per row, and $x^{\mathrm{R}}$ the number of module rows.}
    \label{fig:battery_hierarchy}
\end{figure*}

The decision vector is

\begin{equation}
\mathbf{x}
=
\left[
x^{\mathrm{C}},
x^{\mathrm{T}},
x^\mathrm{S},
x^{\mathrm{M}},
x^\mathrm{R}
\right]^\top,
\end{equation}

\noindent where $x^{\mathrm{C}}$ denotes cathode chemistry, $x^\mathrm{T}$ positive-electrode thickness, $x^{\mathrm{S}}$ the number of cells per module, $x^{\mathrm{M}}$ the number of modules per row, and $x^{\mathrm{R}}$ the number of module rows. We keep the BatPaC defaults of four cells connected in parallel within a module and vary the number of cells per module through $x^{\mathrm{S}}$. The last four variables are the same as in Song et al.~\cite{song2026dynamic}, who adopted them from earlier physics-based battery design optimization studies \cite{sakti2015battery,farkhandi2023batteryoptimization}. Figure~\ref{fig:battery_hierarchy} shows how they map onto the cell, module, and pack hierarchy of BatPaC. We add cathode chemistry as a design variable since manufacturers choose it by weighing material cost against energy, power, and pack size, so a change in the design condition can change which chemistry is preferable. We consider lithium iron phosphate (LFP) and lithium nickel manganese cobalt oxide (NMC811), the two cathode chemistries most widely used in current automotive cells \cite{link2023batterycell,iea2025gevo}. Table~\ref{tab:design_domains} gives the variable types and bounds of the design domain \(\mathcal{X}\). Because \(\mathcal{X}\) combines categorical, continuous, and integer variables and BatPaC is an external simulator, the problem is a simulator-driven mixed-integer nonlinear optimization problem (MINLP).

\begin{table}[t]
\caption{Battery-pack design variables used in the redesign study}
\label{tab:design_domains}
\centering
\begin{tabular}{llll}
\toprule
Var & Description & Type & Domain \\
\midrule
$x^{\mathrm{C}}$ & Cathode chemistry & categorical & $\{\mathrm{LFP},\mathrm{NMC811}\}$ \\
$x^{\text{T}}$ & Thickness (LFP) & continuous & $[50,165]~\mu\mathrm{m}$ \\
 & \quad (NMC811)&  & $[50,155]~\mu\mathrm{m}$\\
$x^{\text{S}}$ & Cells per module & integer & $[1,30]$ \\
$x^{\mathrm{M}}$ & Modules per row & integer & $[1,50]$ \\
$x^{\text{R}}$ & Rows per pack & integer & $[1,4]$ \\
\bottomrule
\end{tabular}
\end{table}

Cathode thickness requires an additional practical bound because increasing electrode thickness creates competing effects in battery design. A thicker electrode increases active-material loading and can improve energy capacity per unit volume or mass, but increasingly thick electrodes also experience stronger transport limitations and polarization at high rates \cite{song2026ecs}, reducing maximum power draw per unit energy capacity. Following earlier battery design optimization work and empirical automotive cell data, which show LFP cells using thicker positive electrodes than nickel-rich cells \cite{sakti2015battery,link2023batterycell}, we use chemistry-specific practical upper bounds of 155~$\mu$m for NMC811 and 165~$\mu$m for LFP.

Let \(\boldsymbol{\theta}\) denote the design condition given to the battery design problem, with \(\boldsymbol{\theta}_0\) denoting the common reference condition. As in Song et al.~\cite{song2026dynamic}, the objective is the BatPaC pack cost per usable energy, \(C(\mathbf{x};\boldsymbol{\theta})\) in USD/kWh, and the constraints bound the energy $E$, power $P$, pack voltage $V$, cell capacity $c$, pack length $L_{\mathrm{BAT}}$, and pack width $W_{\mathrm{BAT}}$ that BatPaC computes from the design. %
Under the common reference condition, the optimization problem is defined as

\begin{equation}
\begin{aligned}
\min_{\mathbf{x}\in\mathcal{X}} \quad
& C(\mathbf{x};\boldsymbol{\theta}_0) \\
\mathrm{s.t.} \quad
& E(\mathbf{x}) \ge 84~\mathrm{kWh}, \quad P(\mathbf{x}) \ge 239~\mathrm{kW}, \\
& 200~\mathrm{V} \le V(\mathbf{x}) \le 1000~\mathrm{V}, \quad 40~\mathrm{Ah} \le c(\mathbf{x}) \le 120~\mathrm{Ah}, \\
& L_{\mathrm{BAT}}(\mathbf{x}) \le 2200~\mathrm{mm}, \quad W_{\mathrm{BAT}}(\mathbf{x}) \le 1500~\mathrm{mm}.
\end{aligned}
\label{eq:battery_source_problem}
\end{equation}

The reference energy and power targets are selected to represent a contemporary long-range electric crossover vehicle. In particular, the Hyundai IONIQ~5 Long Range AWD uses an 84~kWh battery and has a combined traction-motor output of 239~kW \cite{hyundai2026ioniq5spec}. We use these two values as the energy and power requirements in Eq.~\eqref{eq:battery_source_problem}.

The remaining bounds are set so that the domain covers practical automotive conditions. The 200--1000~V voltage range is taken from prior simulator-driven battery optimization work \cite{song2026dynamic} and covers both conventional approximately 400-V architectures and newer 800-V architectures \cite{jung2017powerup,farkhandi2023batteryoptimization}. The 40--120~Ah cell-capacity range is also consistent with previous BatPaC-based design work and reported pouch and prismatic cell capacities \cite{farkhandi2023batteryoptimization,link2023batterycell}. The 2200~mm by 1500~mm packaging envelope represents the scale of full-floor EV battery packs \cite{seresBatteryPack,teslaModelSPackLBNL}.

The common reference design is obtained from a deterministic preliminary grid search under the reference condition defined in Eq.~\eqref{eq:battery_source_problem}. The search uses a discrete mesh over cathode chemistry, cathode thickness, cells per module, modules per row, and module rows. Appendix~\ref{app:source_grid} gives the details of the search. Among the evaluated candidates satisfying the reference constraints, the lowest-cost grid point is

\begin{equation}
\mathbf{x}^{\mathrm{REF}}
=
[\mathrm{NMC811},155~\mu\mathrm{m},29,2,4]^\top.
\label{eq:source_design}
\end{equation}

Under the reference condition, this design evaluates to 96.5~USD/kWh with $E=84$~kWh, $P=239$~kW, $V=202$~V, $c=104$~Ah, $L_{\mathrm{BAT}}=1260$~mm, and $W_{\mathrm{BAT}}=1400$~mm. The same reference design is used for every redesign problem and serves as the best-known solution of the reference problem.

\subsection{Redesign Scenarios}
\label{sec:scenarios}

The four redesign scenarios are denoted S1--S4 for the material-cost, space-constrained, fast-charge, and vehicle-energy cases, respectively. Table~\ref{tab:redesign_scenarios} summarizes their numerical changes from the reference condition, where CAM denotes cathode active material. Requirements not listed in the table remain unchanged.

\begin{table}[t]
\caption{Reference-to-redesign changes for the four redesign scenarios}
\label{tab:redesign_scenarios}
\centering
\setlength{\tabcolsep}{4.5pt}
\begin{tabular}{lll}
\toprule
Scenario & Changed quantity & Reference $\rightarrow$ Redesign \\
\midrule
S1 & NMC811 CAM cost
& 19.63 $\rightarrow$ 25.75~USD/kg \\

S2 & Maximum pack width
& 1500 $\rightarrow$ 1100~mm \\

S3 & Recharge time
& 60 $\rightarrow$ 30~min \\

\multirow{3}{*}{S4}
& Required pack energy
& 84 $\rightarrow$ 130~kWh \\

& Maximum pack length
& 2200 $\rightarrow$ 2700~mm \\

& Maximum pack width
& 1500 $\rightarrow$ 1650~mm \\
\bottomrule
\end{tabular}
\end{table}
\textbf{S1: Change in material prices.} Cobalt production is concentrated in the Democratic Republic of the Congo, so cobalt-bearing cathode materials are exposed to supply and price shocks \cite{nncta2023criticaltechnology,oconnor2026supply,cheng2024supplychain}. S1 represents such a disruption, which increases the cost of cobalt-containing NMC811 cathode material. To translate this disruption into a cathode active material cost, we use the bottom-up CAM-cost framework of Wentker et al.~\cite{wentker2019camcost}. 
Using a cobalt disruption condition informed by the scenarios of O'Connor et al.~\cite{oconnor2026supply}, this calculation gives NMC811 CAM costs of 26.5 and 34.8~USD/kWh under the baseline and disrupted conditions, respectively. We transfer the resulting relative cost increase to the NMC811 CAM input used in the present BatPaC model,
\begin{equation}
p_{\mathrm{NMC}}^{\mathrm{TAR}}
=
19.63
\left(
\frac{34.79}{26.52}
\right)
=
25.75~\mathrm{USD/kg}.
\label{eq:nmc_cost_shock}
\end{equation}

The LFP CAM cost remains unchanged at 8.70~USD/kg because the cobalt disruption does not directly affect the cobalt-free LFP chemistry. This scenario changes the objective while leaving the common feasibility requirements unchanged, so the reference design remains feasible but may no longer be cost competitive.

\textbf{S2: Change in space constraints. }S2 represents redesign caused by a reduction in the physical space allocated to the battery within the vehicle. Vehicle programs also develop many subsystems in parallel, so a revision elsewhere in the vehicle, such as a change to the door or body structure, can reduce the space left for the pack after the battery layout has already been set \cite{widyantara2022batterypacking}. Following the 3beLiEVe project, which sets a maximum battery-pack size of $1500\times1100\times150$~mm for segment A/B battery-electric vehicles \cite{3believe2021requirements}, we reduce the allowable pack width from 1500 to 1100~mm. Because the reference design has a width of 1400~mm, this change directly invalidates the existing layout and requires a geometry-driven redesign of the module and pack topology.

\textbf{S3: Change in charging requirements.} S3 introduces a faster charging requirement that activates additional thickness-dependent feasibility in BatPaC. Charging time is a purchase-relevant vehicle attribute, and shortening it constrains cell design because the achievable rate depends on transport and polarization within the electrode \cite{weiss2021fastcharging}. The reference condition uses a 60-min recharge requirement, and the target condition reduces it to 30~min as in prior BatPaC-based battery design studies \cite{farkhandi2023batteryoptimization}. For the target problem, BatPaC computes a candidate-specific maximum allowable cathode thickness. The simulator returns the feasibility outcome together with the margin to that limit, and this margin enters the penalized objective as a constraint violation.

\textbf{S4: Change in energy requirements. }S4 represents redesign for a vehicle class requiring substantially greater battery energy. The target energy is increased from 84 to 130~kWh, which moves the problem from the scale of the IONIQ~5 reference vehicle toward that of the extended-range Ford F-150 Lightning \cite{ford2025lightning}. Because a larger battery also requires additional packaging space, the maximum pack length and width are relaxed to 2700 and 1650~mm, respectively. Reusing the reference topology at the 130-kWh target produces a cell capacity of approximately 161~Ah, which exceeds the 120-Ah limit. The target problem therefore raises cell capacity while also requiring a redistribution of the pack topology.

We ensured that every target problem contains feasible designs by evaluating the preliminary grid of Appendix~\ref{app:source_grid} under each target condition, reported in Appendix~\ref{app:extended}. These reference evaluations were used only for problem construction and subsequent interpretation and were not provided to the language models or the numerical optimizers during the main experiments.
\section{Methods}
\label{sec:methods}

This section defines the redesign formulation and experimental methodology used to evaluate LLM-assisted reoptimization. We first formalize the engineering redesign problem as a reoptimization task that starts from the best-known reference design. We then describe the experimental design, LLM configurations and integration strategies, numerical search methods, and procedures used to evaluate redesign performance.

\subsection{Engineering Redesign}

We formalize engineering redesign as reoptimization of an existing design under a changed design condition. The reference problem, subject to a vector of constraints $\mathbf{g}(\mathbf{x};\boldsymbol{\theta}_0)\le \mathbf{0}$, is

\begin{equation}
\begin{aligned}
\min_{\mathbf{x}\in\mathcal{X}} \quad
& f(\mathbf{x};\boldsymbol{\theta}_0) \\
\mathrm{s.t.} \quad
& \mathbf{g}(\mathbf{x};\boldsymbol{\theta}_0)\le \mathbf{0}
\end{aligned}
\label{eq:source_problem}
\end{equation}

\noindent with its best-known solution $\mathbf{x}^{\mathrm{REF}}$, which we call the reference design. When the design condition changes from $\boldsymbol{\theta}_0$ to $\boldsymbol{\theta}_r$, the existing design must be reconsidered under the redesign problem

\begin{equation}
\begin{aligned}
\min_{\mathbf{x}\in\mathcal{X}} \quad
& f(\mathbf{x};\boldsymbol{\theta}_r) \\
\mathrm{s.t.} \quad
& \mathbf{g}(\mathbf{x};\boldsymbol{\theta}_r)\le \mathbf{0}
\end{aligned}
\label{eq:redesign_problem}
\end{equation}

\noindent In all redesign experiments, the redesign formulation and $\mathbf{x}^{\mathrm{REF}}$ are available to the redesign procedure.

\subsection{Experimental Design}

Figure~\ref{fig_framework} summarizes the experimental design, and the panels correspond to the choices that define one run. The panels outlined in green contain the experimental factors (the redesign scenario, the semantic-information level, and the language model size). The two numbered groups contain the redesign methods, and the numerical optimizer is shown between them. Without a language model, such a redesign is handled by a conventional numerical optimizer started either from randomly sampled designs or from the existing design. We call these the cold start and the reference-warm baseline, and use them as references. WAFER replaces that step with one of three language model roles (warm starting, variable fixing, and ReAct). Warm starting and variable fixing still use one of four downstream optimizers for the remaining search, so a run using them also selects an optimizer, whereas ReAct proposes every candidate itself and does not use one. The gray panels contain the design problem and the BatPaC simulator, which every run shares. The arrows show the information passed between them. Sections~\ref{sec:semantic} to \ref{sec:outcomes} define each of these elements.

All methods operate under a common simulation budget of $B=20$ accepted BatPaC evaluations per run. An evaluation is accepted when the proposed design passes validation and is evaluated by BatPaC, and rejected proposals do not consume the budget, as Appendix~\ref{app:prompts} describes. Because simulator wall time and cost can vary substantially across engineering problems and simulation tools, we use simulator evaluations as a common measure of engineering search effort.

The number of configurations differs by method since the methods use different components. The cold start and the reference-warm baseline vary over four scenarios and four downstream optimizers, which gives 16 configurations each. Warm starting and variable fixing vary over the same scenarios and optimizers and additionally over three model sizes and two semantic levels, which gives 96 configurations each. ReAct proposes every candidate itself and therefore has no downstream optimizer, which gives 24 configurations. The full design contains 248 configurations, and every configuration contains 50 runs, for a total of 12,400 runs.

\subsection{Semantic Information and Language Models}
\label{sec:semantic}

The semantic-information treatment controls how much engineering knowledge is provided beyond the numerical optimization formulation. In L1, the model receives the design variables and bounds, reference design, redesign requirements, objective, and constraint definitions. In L2, this information is supplemented with compact engineering relationships relevant to battery redesign, including chemistry-dependent cost and energy tradeoffs, electrode-thickness effects, voltage and current relationships, and geometric consequences of pack topology. We wrote these relationships from the BatPaC manual \cite{knehr2022batpac} and basic electrochemical principles so that they cover the changes in the four redesign scenarios. Both levels are placed in the user message that accompanies a fixed system message. The comparison therefore tests whether explicit engineering relationships provide useful guidance beyond the problem definition alone. Appendix~\ref{app:prompts} reproduces the complete prompts.

We use Qwen3.5 as the language model, at the 4B, 9B, and 27B parameter sizes with the released bfloat16 weights and no quantization, and we do not perform any task-specific fine-tuning \cite{qwen2026qwen35}. Qwen3.5 supports explicit reasoning and tool calling,
which are needed to interpret engineering information and interact with the external simulator during redesign. The publicly available model weights also allow local deployment, providing control over model scale and inference infrastructure while avoiding the need to transmit potentially sensitive engineering information to a hosted model service. Keeping the same model family and prompting framework across the three sizes lets us attribute the difference to model scale. Model outputs are converted to structured redesign actions and validated before simulator evaluation. Exact prompts, sampling settings, and output schemas are provided in Appendix~\ref{app:prompts}.

Because the L2 treatment relies on the model correctly interpreting the supplied engineering information, we separately tested its comprehensibility using Qwen3.5-4B, the smallest model considered. Across 12 tests spanning all four redesign scenarios, the model recovered and interpreted the main L2 engineering relationships in nearly all cases and consistently distinguished quantities that require BatPaC evaluation from information available in the prompt. Appendix~\ref{app:l2_comprehension} discusses the test in detail.

\subsection{LLM Integration Strategies}
\label{sec:strategies}

The three LLM strategies represent different ways in which semantic information can influence redesign. Warm starting (WS) uses the LLM before numerical optimization to propose three initial designs. These designs are evaluated by BatPaC and used to initialize the downstream optimizer, testing whether semantic guidance can provide starting points that either lie in promising regions of the redesigned search space or help the optimizer learn the behavior of the objective.

Variable fixing (VF) instead uses the LLM to determine which variables of the reference design should remain fixed and which should be released for optimization. The downstream optimizer then searches only the variables that remain free, starting from a random point in that reduced space. This strategy tests whether engineering knowledge can reduce the effective search dimension by preserving design decisions that remain useful after the redesign condition changes, and whether that reduction improves optimization performance.

ReAct uses semantic guidance throughout the search and follows the reasoning-and-acting loop introduced for language model agents \cite{yao2023react}, in which reasoning steps and actions alternate. The model states its reasoning, proposes a design, and receives the objective and constraint values of the design and its feasibility from BatPaC as the observation for that step. The action and observation of every earlier step are carried forward in the prompt, so each new candidate is chosen based on the accumulated record. ReAct therefore tests whether engineering judgment based on the trajectory of evaluated designs and simulator feedback improves the search at each step.

\subsection{Numerical Search and Reference Conditions}

Black-box optimizers choose the next design from function evaluations alone. They are commonly divided into model-based methods, which fit a surrogate of the objective, and direct-search methods, which compare function values directly \cite{rios2013dfo,larson2019dfo}. For expensive black-box problems, Bayesian optimization and evolutionary search are regarded as the two principal families \cite{gobert2024bayesianevolutionary}. We therefore pair WS and VF with four methods that cover both classifications: Bayesian optimization (BO) as a model-based method, a mixed-variable genetic algorithm (GA) as an evolutionary method, mesh adaptive direct search (MADS) as a direct-search method, and random search as a nonadaptive reference that uses no feedback \cite{bergstra2012random}. BO uses a probabilistic surrogate model and an acquisition function to guide evaluations of expensive black-box objectives \cite{jones1998ego,frazier2018bo}. We implement it with the Ax adaptive experimentation framework, which supports mixed continuous and discrete search spaces \cite{olson2025ax}. The GA evolves a population of designs by recombining promising candidates and mutating them to explore nearby designs. Because the battery problem combines categorical, continuous, and integer variables, we implement the GA with the mixed-variable algorithm of pymoo \cite{blank2020pymoo}. MADS is a model-free direct-search approach developed for constrained black-box optimization and widely applied to simulation-based engineering problems \cite{audet2006mads,audet2022nomad4}.

All four optimizers receive the same penalized scalar objective. When a design is feasible, the objective is its BatPaC cost. Otherwise, a fixed penalty and a term proportional to the total normalized constraint violation are added to the cost. This keeps the information available to the four optimizers identical. We additionally evaluate non-LLM numerical baselines using each of the same four optimization methods. 

In the cold numerical condition, the optimizer begins from designs sampled uniformly at random from the design domain without information from the reference design. In the reference-warm condition, the reference design $\mathbf{x}^{\mathrm{REF}}$ is included in the numerical initialization. Appendix~\ref{app:optimizer_settings} gives the exact penalty, the optimizer settings, the initialization procedures, and the software configurations.

\subsection{Evaluation and Statistical Analysis}
\label{sec:outcomes}

Because ReAct carries its interaction history in the prompt, its context grows with every accepted evaluation and with every rejected proposal. We limited the budget to 20 evaluations so that this history stays within the 32,768-token context window of the three model sizes, and no run in this study terminated because of context length.

We evaluate redesign effectiveness by how often and how early a method finds a feasible design and by how low a cost it reaches. Feasible-design discovery records whether a method identifies at least one design satisfying all redesign constraints. Search efficiency is measured by the accepted evaluation index of the first feasible design. Design quality is measured by the lowest feasible BatPaC cost obtained within the 20-evaluation budget. We also measure computational effort by LLM inference time and token use, and we report the number of runs that terminate without producing a usable decision.

Neither the numerical optimizers nor the language model is deterministic, so a single run of a configuration provides noisy information about that configuration. We therefore repeat every configuration over 50 seeds and compare distributions of our performance metrics.

We compare feasible discovery and near-best attainment between configurations with a two-sided Fisher exact test \cite{fisher1935design} and adjust the $p$-values with the Holm procedure within each family defined by one scenario, one strategy, and one metric \cite{holm1979simple}. Section~\ref{sec:res_overall} states how the configurations are paired. Proportions shown in the figures are reported with 95\% Wilson intervals \cite{wilson1927probable}. The association between the VF release and WS move probabilities in Sec.~\ref{sec:res_mechanism} is measured with the Spearman rank correlation \cite{spearman1904proof}. For proportions pooled over configurations, the Wilson intervals reflect sampling variation among runs and do not include differences between configurations.

\section{Results}
\label{sec:results}

This section compares the LLM-assisted redesign strategies proposed in this study with the conventional numerical optimizers that would be used without an LLM. We first report quantitative performance in Sec.~\ref{sec:res_overall}. We then analyze the logic behind why each redesign method produces better performance in Sec.~\ref{sec:res_mechanism}. This analysis gives intuition about which method is appropriate for each type of redesign scenario. We finally analyze how the language model scale affects redesign performance and inference cost in Sec.~\ref{sec:res_scale}.

The four redesign scenarios S1 to S4 are separate optimization problems with structurally different redesign tasks, so we interpret performance separately for each scenario. We define a configuration as one combination of scenario, redesign method, model size, semantic level, and downstream optimizer. Throughout this section the cold start is denoted Cold, the reference-warm numerical baseline is denoted RW, and the three WAFER methods are denoted VF for variable fixing, WS for warm starting, and ReAct. The result figures use the method colors of Fig.~\ref{fig_framework}, so VF is green, WS is red, and ReAct is blue, while the numerical baselines are black for Cold and light gray for RW.

\subsection{LLMs can improve feasible discovery and design quality}
\label{sec:res_overall}

We judge each method by whether it finds a design that satisfies the changed constraints, how quickly it does so, and how far it then improves that design under the objective.

\begin{figure}[t]
    \centering
    \includegraphics[width=\linewidth]{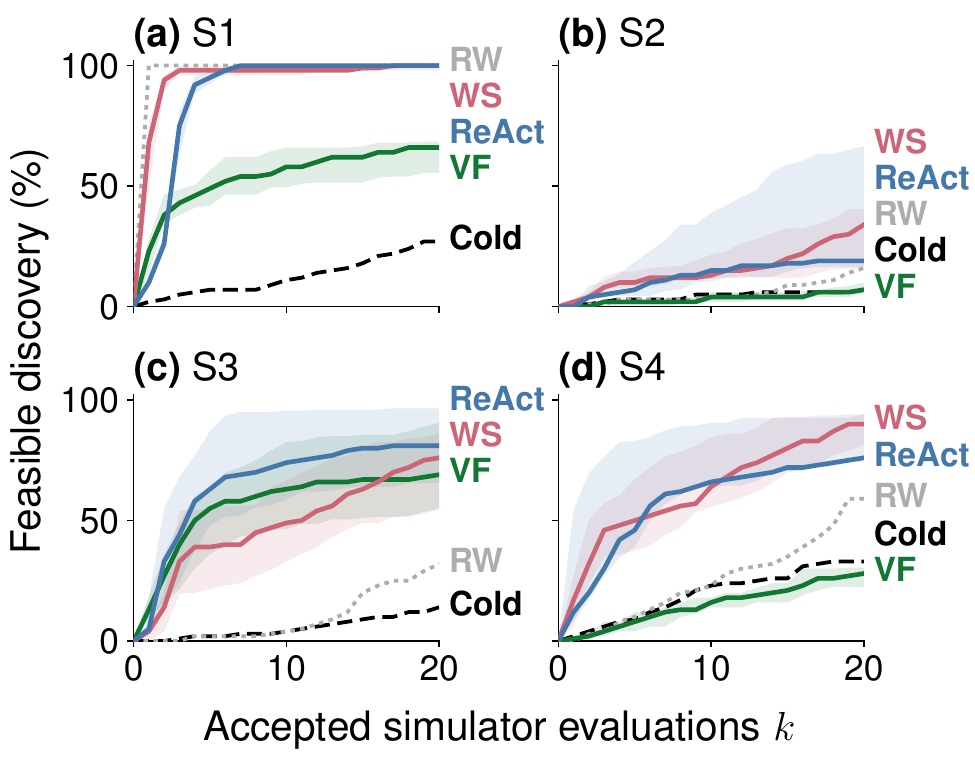}
    \caption{Cumulative feasible discovery over 20 accepted simulator evaluations. Curves show the median across configurations of the fraction of each configuration's 50 runs that have found a feasible design, and shaded bands show the interquartile range across configurations. RW denotes the reference-warm baseline.}
    \label{fig:R1a}
\end{figure}

Figure~\ref{fig:R1a} shows cumulative feasible discovery against the number of accepted evaluations $k$. For each configuration we first compute the fraction of its 50 runs that have found a feasible design, which gives one value per configuration at every $k$. The curve is the pointwise median of these configuration values and the shaded band is their interquartile range. The figure pools model size, semantic level, and downstream optimizer, so only the effect of the redesign method remains. A curve that lies higher means that a larger fraction of runs has found a feasible design after the same number of evaluations. Below, we also utilize a single number pooled over all runs with respect to each reoptimization method.

In three of the four scenarios, the curves for WS and ReAct lie clearly above the curves for the numerical baselines (S1, where the reference design is feasible, is the exception, since RW begins feasible). S3 shows the separation most clearly: Feasible discovery at the end of the budget is 77\% for ReAct and 69\% for WS, compared with 35\% for RW and 17\% for the cold start. Counting runs without a feasible design as exceeding the budget, the median index of the first feasible design in S3 is 4 for ReAct and 11 for WS, whereas more than half of the RW and cold-start runs found no feasible design within 20 evaluations. The same ordering holds in S4, where the median is 4 for both WS and ReAct against 17 for RW, and more than half of the cold-start runs found none. In S2, more than half of the runs of every method found no feasible design, but WS and ReAct still reach feasibility more often than the two baselines.

S1 behaves differently because its reference design is already feasible. A method therefore achieves feasible discovery by submitting the reference design without modification, and the RW curve reaches 100\% at the first evaluation. WS at 99\% and ReAct at 100\% saturate almost as quickly, which shows that inserting a language model does not interfere with recovering a design that is already feasible.

VF reaches 72\% in S3, which is comparable to the other LLM-assisted methods, but it is not efficient at finding feasible solutions in the remaining scenarios. The clearest case is S4, where feasible discovery for VF is 27\% against 61\% for RW and 37\% even for the cold start, so variable fixing there is worse than starting with no information about the reference design.

Figure~\ref{fig:R1a} also indicates the relative difficulty of the four redesign scenarios. S2 is the most difficult, because the curves of every method stay near the bottom of the panel. S1 is the easiest, because its reference design is already feasible. S3 is comparatively easy for the LLM-assisted methods and difficult for the numerical baselines, because the curves separate strongly between the two groups.

\begin{figure}[t]
    \centering
    \includegraphics[width=\linewidth]{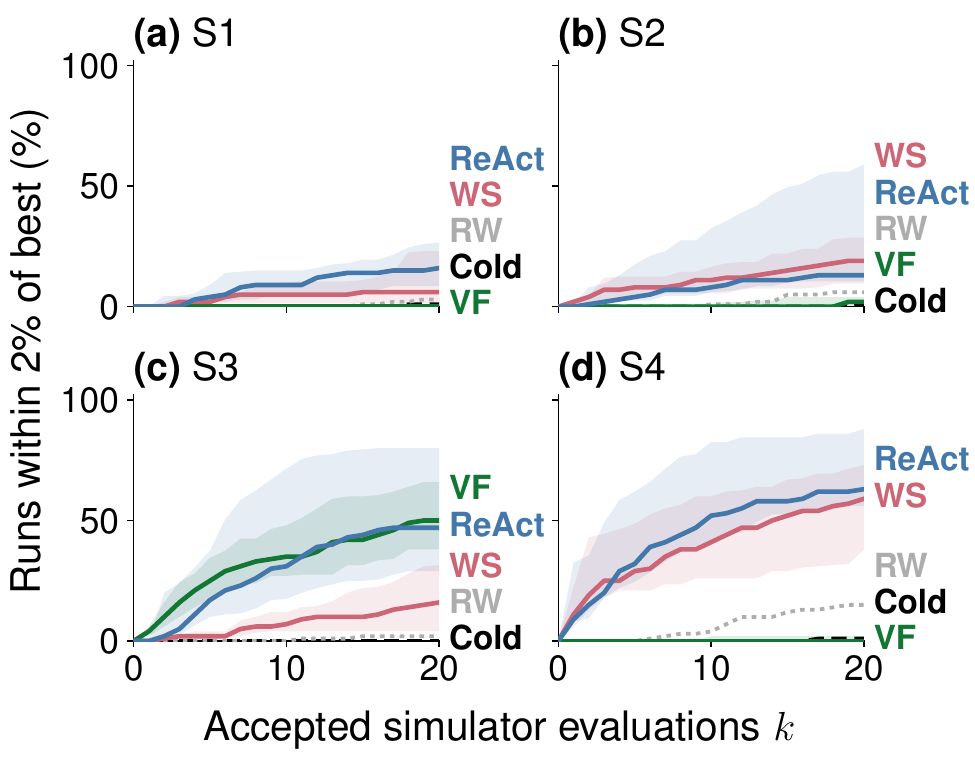}
    \caption{Cumulative attainment of an optimality gap of 2\% or less over 20 accepted simulator evaluations. The fraction is computed over all runs. The best-known is the best feasible cost observed for each redesign problem. Curves and bands are defined as in Fig.~\ref{fig:R1a}.}
    \label{fig:R1b}
\end{figure}

Table~\ref{tab:bestdesign} reports the best feasible design found across all runs in each redesign problem. The cobalt price shock in S1 makes the cobalt-free chemistry preferable, so the best design switches from NMC811 to LFP at its thickness bound. The reduced pack width in S2 keeps the chemistry and electrode thickness and rearranges the pack to three modules per row and three rows with three fewer cells per module. The fast-charge requirement in S3 keeps the topology and reduces the cathode thickness from 155 to 97~$\mu$m. The larger energy target in S4 combines both kinds of change, adopting LFP at its thickness bound and adding one module per row.

In each scenario we take the cost of the best design found across all runs as the best-known cost, which are 97.1~USD/kWh for S1, 97.4~USD/kWh for S2, 100~USD/kWh for S3, and 90.3~USD/kWh for S4, and we use them to analyze how closely each method approaches the best known solution. Figure~\ref{fig:R1b} shows the fraction of runs that reached an optimality gap of 2\% or less, which is a cost within 1.02 times the best-known cost, while satisfying all constraints. The horizontal axis, curves, and bands are the same as in Fig.~\ref{fig:R1a}. The fraction is computed over all runs, including those that never reached feasibility.

\begin{table}[t]
\caption{Best feasible design found in each redesign problem, against the common reference design. The cost of each redesign row is the best-known cost used for the optimality gap.}
\label{tab:bestdesign}
\centering
\setlength{\tabcolsep}{4pt}
\begin{tabular}{lllrrrr}
\toprule
Sc. & $x^{\mathrm{C}}$ & $x^{\mathrm{T}}$ ($\mu$m) & $x^{\mathrm{S}}$ & $x^{\mathrm{M}}$ & $x^{\mathrm{R}}$ & \shortstack{Cost\\(USD/kWh)} \\
\midrule
Reference & NMC811 & 155 & 29 & 2 & 4 & 96.5 \\
\midrule
S1 & LFP    & 165 & 28 & 3 & 3 & 97.1  \\
S2 & NMC811 & 155 & 26 & 3 & 3 & 97.4  \\
S3 & NMC811 & 96.7  & 29 & 2 & 4 & 100 \\
S4 & LFP    & 165 & 28 & 3 & 4 & 90.3  \\
\bottomrule
\end{tabular}
\end{table}

ReAct and WS again reach the near-best region more often than the numerical baselines in all four scenarios. The cold start stayed at or below 1\% in every scenario. VF is again the exception to the ordering. It rarely reaches the near-best region in S1, S2, and S4, but it reaches 53\% in S3, which is close to the 54\% of ReAct and well above the 19\% of WS. S1 shows the complementary case. It was the easiest scenario for feasibility, but the curves of every method stay low in Fig.~\ref{fig:R1b}, which shows that the two criteria differ in difficulty.

Taken together, WS and ReAct solve the redesign optimization problem better than the conventional numerical methods in general, and VF is dominant only in S3, where the shorter recharge time requires a change in the cathode thickness. Section~\ref{sec:res_vf} explains why this structure favors VF. Pooled over the four scenarios, the LLM-assisted runs reach after four evaluations the feasible-discovery rate of 39\% that the numerical baselines reach after 20, and 23\% of them end within 2\% of the best-known cost against 7\% for the numerical baselines.

\begin{table}[t]
\caption{Optimizer-matched comparisons with the reference-warm baseline at the end of the budget. Each entry is the number of comparisons in which the method is significantly better than RW, followed by the number in which it is significantly worse, under the Holm-adjusted two-sided Fisher exact test at $\alpha=0.05$. Each VF or WS configuration is compared with the RW configuration that uses the same downstream optimizer, and each of the six ReAct configurations is compared with all four RW configurations, which gives 24 comparisons per entry. Disc.\ is feasible discovery and Gap is attainment of an optimality gap of 2\% or less. Bold marks a nonzero count of significantly better comparisons.}
\label{tab:strategy}
\centering
\begin{tabular}{lcccccc}
\toprule
 & \multicolumn{2}{c}{VF} & \multicolumn{2}{c}{WS} & \multicolumn{2}{c}{ReAct} \\
\cmidrule(lr){2-3}\cmidrule(lr){4-5}\cmidrule(lr){6-7}
Sc. & Disc. & Gap & Disc. & Gap & Disc. & Gap \\
\midrule
S1 & 0/24 & 0/6 & 0/0 & 0/0 & 0/0 & \textbf{7}/2 \\
S2 & 0/5 & 0/4 & \textbf{1}/0 & \textbf{1}/0 & \textbf{8}/2 & \textbf{8}/4 \\
S3 & \textbf{17}/0 & \textbf{24}/0 & \textbf{18}/0 & \textbf{7}/0 & \textbf{17}/1 & \textbf{20}/0 \\
S4 & 0/14 & 0/6 & \textbf{11}/0 & \textbf{14}/0 & \textbf{10}/0 & \textbf{20}/1 \\
\bottomrule
\end{tabular}
\end{table}

Figures~\ref{fig:R1a} and \ref{fig:R1b} pool all configurations, so they show how a method performs when its setting is chosen without matching approach to scenario. Because the downstream optimizer can change the search behavior substantially, we also test whether the differences between the methods hold when the optimizer is held fixed. Table~\ref{tab:strategy} reports this test. Each VF or WS configuration is compared with the RW configuration that shares its scenario and downstream optimizer, and each of the six ReAct configurations is compared with all four optimizer-specific RW configurations. Table~\ref{tab:full_results} in Appendix~\ref{app:extended} gives the values of every configuration and shades those that are significantly better than RW.

Under matched optimizers, WS and ReAct are significantly better than RW in a large share of the comparisons and are significantly worse in only a few. VF shows the opposite pattern. It is significantly worse than RW in most comparisons of S1 and S4 and it is never significantly better in those two scenarios, while in S3 it is significantly better in every comparison of the 2\% gap and in most comparisons of discovery.

These results raise two questions. How does the language model use the information in the reference design in a way that produces better performance, and why does VF work well only in S3 and not in the remaining scenarios? Section~\ref{sec:res_mechanism} examines these mechanisms and analyzes which method works best in each redesign scenario.

\subsection{The best use of LLMs depends on the redesign scenario}
\label{sec:res_mechanism}

\begin{figure}[t]
    \centering
    \includegraphics[width=\linewidth]{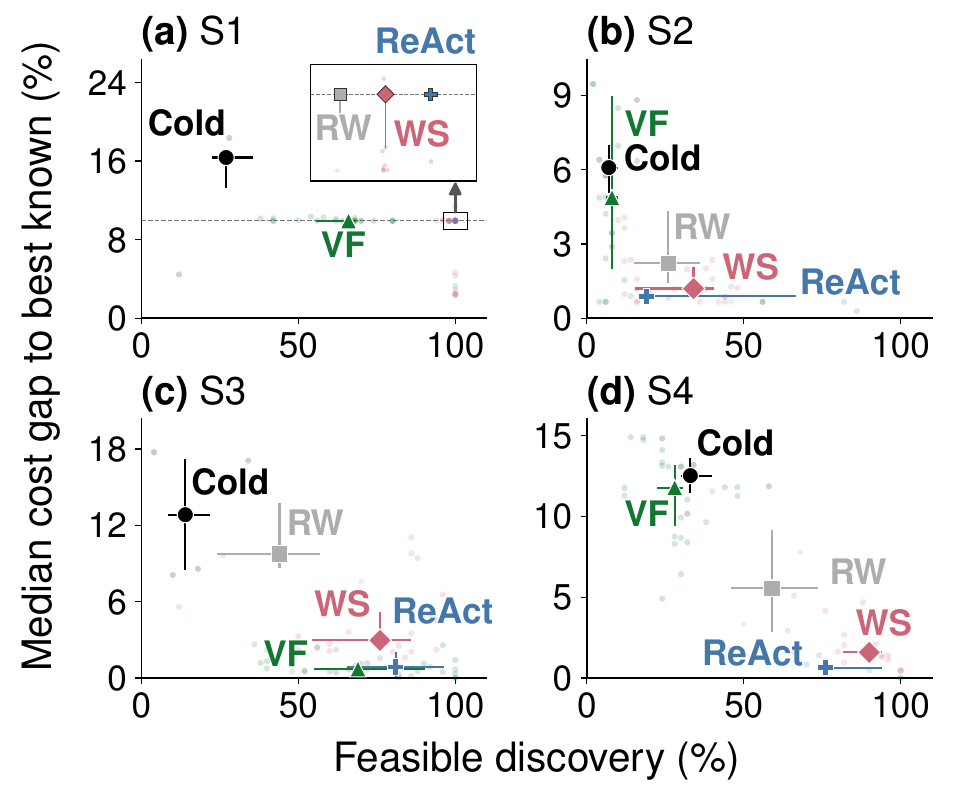}
    \caption{Feasible discovery and the gap of the median final feasible cost to the best-known cost across configurations. The gap is measured against the best-known cost of each scenario in Table~\ref{tab:bestdesign}. Faded points show individual 50-run configurations. Large markers show strategy medians and whiskers show interquartile ranges. Lower-right values designate better optimization performance. The inset enlarges the boxed region in (a). The dashed line in (a) indicates the cost of the reference design. Best outcomes have a low cost gap (down) and high feasible discovery rates (right).}
    \label{fig:R2}
\end{figure}

Figure~\ref{fig:R2} summarizes feasible discovery and the gap of the median final feasible cost to the best-known cost for each configuration. Faded points are individual 50-run configurations, large markers are the medians across the configurations of each method, and whiskers are the corresponding interquartile ranges. The best-known design therefore lies at zero gap. The reference design is already feasible in S1, so we also mark the redesign-condition cost of the reference design at 107~USD/kWh, a gap of 9.9\%, for comparison. We want high feasibility and low cost, so performance improves toward the lower right of each panel.

The best-performing method is an LLM-based method in all cases, but the best use of the LLM differs across scenarios. ReAct and WS perform best in S1, S2, and S4, and VF and ReAct perform best in S3. In S1 the reference design is already feasible, so RW, WS, and ReAct all secure feasibility without difficulty. Lowering the objective is difficult, however, and the medians of the three methods cluster near the reference-design cost, a gap of 9.9\%. The inset of Fig.~\ref{fig:R2}(a) enlarges the boxed region around these medians.

Among the individual runs, the two 27B ReAct configurations, one at each semantic level, take the lower-right corner in S2, S3, and S4. They hold the two highest feasible discovery values in S2 at 86\% and 82\% among the 62 configurations of that panel, and they reach 100\% discovery together with the lowest median final cost in S3 and S4. It is notable that the ReAct formulations do not use numerical solvers and are nevertheless able to compete with or beat approaches using numerical solvers.

The changes in design requirements for each redesign scenario induce necessary changes in design variable values, with the exception of S1, for which the reference design is feasible for the redesign scenario. We examine all feasible designs identified in our preliminary grid search and across all runs of all methods, and we identify variables that change from the reference design values in every known feasible design. We find that these ``required redesign variables'' are $x^{\mathrm{M}}$ and $x^{\mathrm{R}}$ in S2, $x^{\mathrm{T}}$ in S3, and $x^{\mathrm{M}}$ in S4. Because S1 has no required redesign variables, we instead examine $x^\mathrm{C}$ since the best known design and the fifty best feasible designs found from the grid search all involve a change of cathode chemistry from NMC811 to LFP. 
We find that VF and WS reach similar judgments about which variables must change. Across the 20 pairs of scenario and design variable, the probability that VF releases a variable and the probability that WS moves it away from its reference value have a Spearman correlation of $\rho_s=0.79$. The required variables receive higher probabilities under both methods. VF releases them with a mean probability of 85\% against 59\% for the remaining variables, and WS changes them from their reference values with a mean probability of 52\% against 36\%. The two methods therefore agree on which variables must change, and the performance difference arises from how each method uses that judgment in the optimization.

\subsubsection{Variable fixing works best when the necessary redesign involves only a few variables}\label{sec:res_vf}

\begin{figure}[t]
    \centering
    \includegraphics[width=\linewidth]{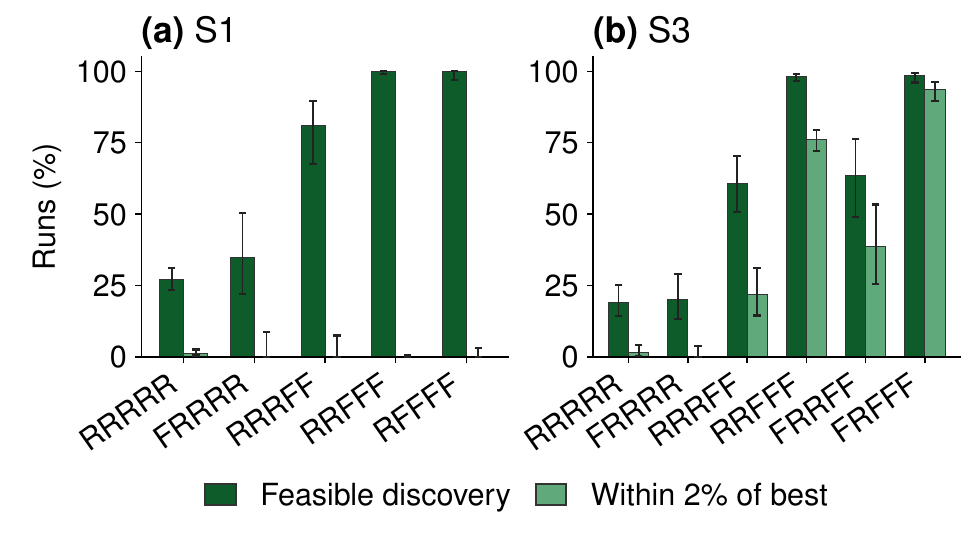}
    \caption{Variable-fixing masks selected in S1 and S3. Bars show feasible discovery and near-best attainment among the runs selecting each mask, and whiskers show 95\% Wilson intervals. Masks are ordered by the number of released variables, and masks selected in fewer than 20 runs are omitted. R and F denote released and fixed variables in the order $(x^{\mathrm{C}},x^{\mathrm{T}},x^{\mathrm{S}},x^{\mathrm{M}},x^{\mathrm{R}})$.}
    \label{fig:R3a}
\end{figure}

The strength of VF is that it reduces the number of variables and thereby makes the downstream optimization more efficient. We write a VF decision as a mask of five letters, one per design variable in the order $(x^{\mathrm{C}},x^{\mathrm{T}},x^{\mathrm{S}},x^{\mathrm{M}},x^{\mathrm{R}})$, where R (released) marks a variable that is released to the optimizer and F (fixed) marks a variable that is fixed at its reference value. RRRRR therefore releases everything and RFFFF releases only the chemistry. Figure~\ref{fig:R3a} shows the outcome of the runs grouped by the mask that the language model selected in S1 and S3. For each mask selected in at least 20 runs, the bars give the fraction of those runs that reached a feasible design and the fraction that reached a gap of 2\% or less, pooled over model sizes and optimizers. The masks are ordered so that fewer variables are released from left to right.

In both scenarios, the fraction of runs that reach feasibility rises sharply as fewer variables are released. The RRRRR mask in S1 releases every variable and only 133 of the 491 runs that selected it reached feasibility, while every one of the 435 runs on the RRFFF mask and every one of the 118 runs on the RFFFF mask reached feasibility. S3 shows the same pattern, where 39 of 204 runs on RRRRR reached feasibility against 499 of 509 runs on RRFFF. In S3 the near-best attainment follows the same ordering, rising from 3 of the 204 runs on RRRRR to 387 of the 509 runs on RRFFF. This suggests that reducing the search space, and thereby making the search more efficient, is the main mechanism by which VF improves performance in the problems where VF works.

VF works well in S3 because the variable that must change is easy to identify. The best S3 design differs from the reference design only in $x^{\mathrm{T}}$, and the shorter recharge time constrains the electrode thickness directly. Every model size released $x^{\mathrm{T}}$ in 100\% of its valid S3 decisions. However, in S2 and S4, identifying the required variables depends on the electrode area together with geometric quantities such as $x^{\mathrm{M}}$ and $x^{\mathrm{R}}$. The electrode area is determined by the nonlinear electrochemical calculation inside BatPaC. These quantities are directly connected to the design condition change in those two scenarios, which makes the identification problem much harder. In both of them, every run that fixed a required variable failed to reach feasibility, since an incorrect fixing decision leaves no feasible design in the remaining subspace and the downstream optimizer searches only that subspace. Table~\ref{tab:si_vf24}(a) in Appendix~\ref{app:extended} reports the mask analysis for S2 and S4.

S2 is difficult even when the correct variables are released, because its feasible region is narrow. Among the masks that fix at least one variable, the one selected most often in S2 is FFFRR, which releases only $x^{\mathrm{M}}$ and $x^{\mathrm{R}}$. Inside that subspace only the cell at $x^{\mathrm{R}}=3$ and $x^{\mathrm{M}}=3$ contains feasible evaluations, and all 50 evaluations in that cell are feasible. FFFRR is nevertheless the second-best of the available S2 masks, with a run-level feasibility of 21\%. In S4 the most frequent mask FRRRR reaches 42\%, and its feasible evaluations concentrate at $x^{\mathrm{R}}=4$, with the highest feasible fraction of 68\% at $x^{\mathrm{M}}=4$. However, feasible discovery inside these narrow subspaces also depends on the downstream optimizer. In the FFFRR subspace of S2, MADS reaches 69\%, against 7\% for GA, the second-best optimizer. Figure~\ref{fig:R3b} and Table~\ref{tab:si_vf24}(b) in Appendix~\ref{app:extended} give the details.

Taken together, VF shows a clear advantage by shrinking the search space when the variables that must change are easy to identify. When it is unclear which variables should be left open, an incorrectly fixed variable removes feasibility from the remaining subspace and the downstream optimizer cannot recover it. However, even when the feasible domain is narrow and complex, VF can still perform well when it is paired with a suitable optimizer such as MADS. It is important to note that, had more iterations been permitted, more released variables may have allowed the solver to reach a better solution. With a limited iteration budget, variable fixing can help use the limited available compute to concentrate search effort in a subspace that can produce greater benefits per iteration.

\subsubsection{Warm starting works best when initial points focus on promising redesign directions}

When the language model chooses the warm-start points, it can either concentrate them in promising directions to reach feasible and near-optimal designs quickly, or spread them over several directions to learn more about the design space. This choice mirrors the trade-off between exploitation and exploration in black-box optimizers such as BO. In the scenarios examined, concentrating on exploitation turned out to be far more efficient.

\begin{figure}[t]
    \centering
    \includegraphics[width=\linewidth]{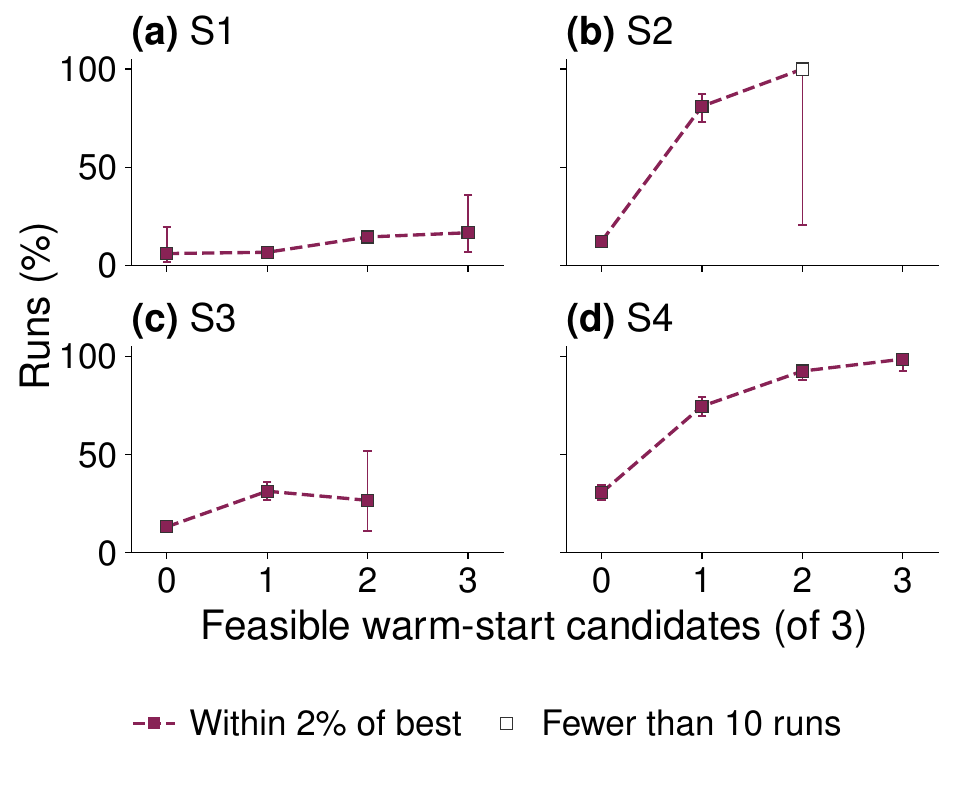}
    \caption{Near-best attainment of warm-start runs versus the number of the three proposed starting points that BatPaC evaluates as feasible. Whiskers show 95\% Wilson intervals, and hollow markers indicate fewer than 10 runs.}
    \label{fig:R4}
\end{figure}

Figure~\ref{fig:R4} groups the WS runs by how many of the three proposed starting points BatPaC evaluates as feasible, and it reports the fraction of runs at a gap of 2\% or less for each group. In S2, S3, and S4, near-best attainment is higher when at least one starting point is feasible, and in S4 it increases with each additional feasible starting point. This suggests that, for WS, placing the starting points in the desired direction is more advantageous than exploring diverse directions.

We also found that when one of the three starting points is the reference design itself, fewer runs reach a gap of 2\% or less in S3 and S4, as Table~\ref{tab:exactsource}(c) in Appendix~\ref{app:extended} shows. Therefore, the limited starting points given to the language model are better spent on regions that the reference design does not already cover. In addition, WS is useful even when the language model knows only the rough direction in which the redesign should search, so it is applicable to many redesign scenarios.

\subsubsection{ReAct is especially useful when objective improvement remains important after feasibility}

ReAct calls the language model at every iteration, so it is the most expensive of the three methods. It also gives the language model full control over the search. Its reasoning trace also allows us to investigate how it balances feasibility and optimality at each step.

\begin{figure}[t]
    \centering
    \includegraphics[width=\linewidth]{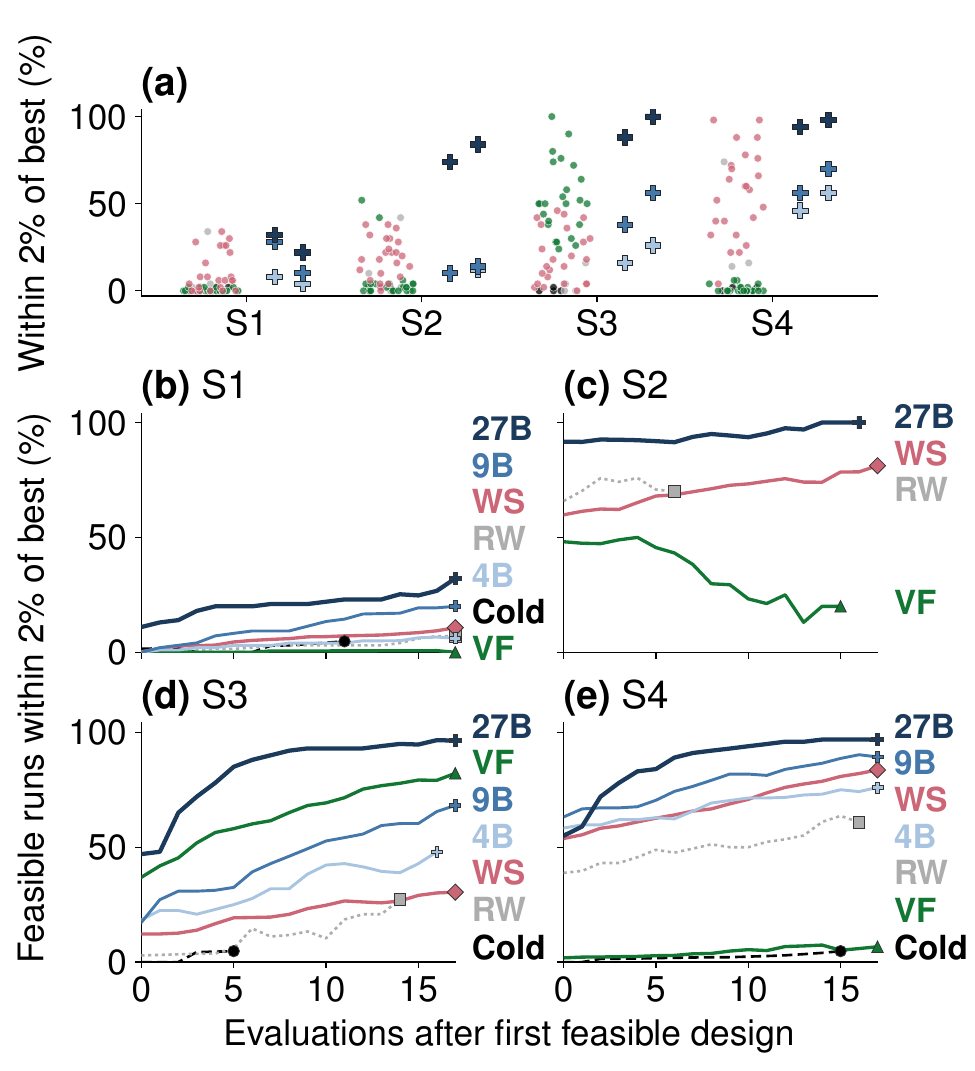}
    \caption{Optimization after the first feasible design. (a) Near-best attainment for each configuration. Plus markers show ReAct configurations by model size, and circles show VF, WS, and numerical configurations. (b)--(e) Fraction of feasible runs reaching an optimality gap of 2\% or less versus accepted evaluations after first feasibility. Curves are shown only while at least 20 runs contribute. In (b) to (e), ReAct curves are labeled by model size and the other curves by method, and (a) uses the same colors.}
    \label{fig:R5}
\end{figure}

Figure~\ref{fig:R5} shows the behavior after the trajectories are aligned at the first feasible design. Panel (a) lists the near-best attainment of every configuration by scenario, where plus markers denote ReAct configurations by model size and circles denote the VF, WS, and numerical configurations. Panels (b) to (e) show the fraction of feasible runs that reach a gap of 2\% or less within $k$ accepted evaluations after the first feasible design, and each curve is drawn only while at least 20 runs remain in the corresponding interval.

The figure shows that ReAct performs best in terms of optimality once a feasible design is found, and that this advantage grows with model size. Among the runs that reach feasibility, the 27B ReAct configurations convert 27\% in S1, 94\% in S2, 94\% in S3, and 96\% in S4 into designs within 2\% of the best-known cost, compared to 14\%, 79\%, 80\%, and 84\% for the best alternative method in each scenario. The 9B model also exceeds the numerical baselines in most scenarios.

Note that the advantage does not come from the larger iteration budget that ReAct has remaining after reaching feasibility early. Among the runs that still had at least ten evaluations remaining, 27B ReAct reaches a gap of 2\% or less within ten evaluations after its first feasible design in 94\%, 93\%, and 94\% of runs in S2 to S4, against at most 83\%, 75\%, and 86\% for the VF and WS configurations in Table~\ref{tab:si_postfeas}(d) in Appendix~\ref{app:extended}.

\begin{figure}[t]
    \centering
    \includegraphics[width=\linewidth]{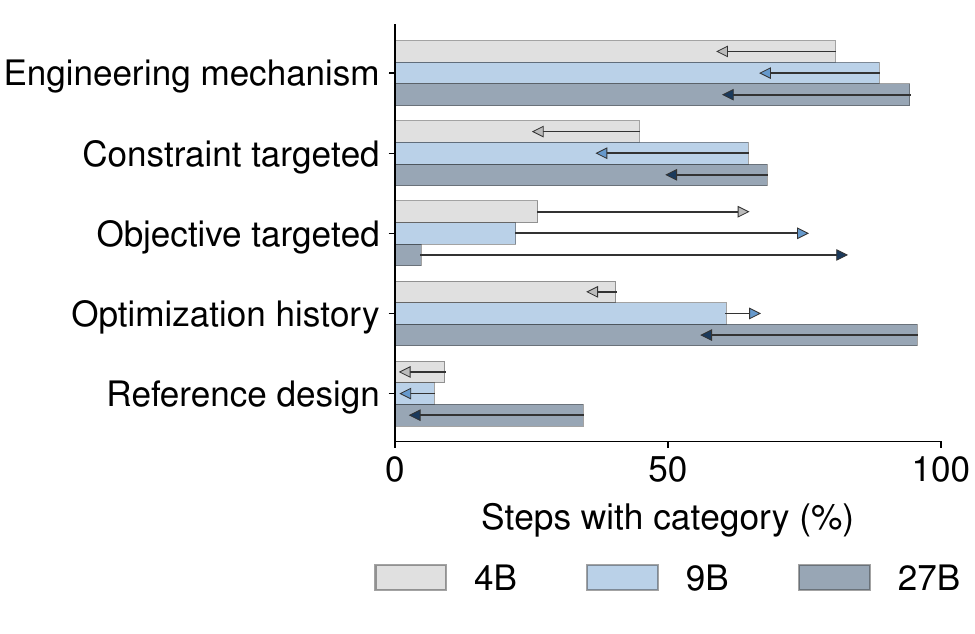}
    \caption{ReAct reasoning before and after feasibility. Bars show the fraction of accepted steps whose reasoning mentions each category, by model size, and arrows run from the value before first feasibility to the value after it.}
    \label{fig:R5b}
\end{figure}

The reasoning trace reflects this tendency. Figure~\ref{fig:R5b} shows the fraction of accepted steps whose reasoning mentions each category before and after the first feasible design. After feasibility, mentions of the objective and cost increase and mentions of constraints decrease for every model size, which indicates that the reasoning moves from satisfying the constraints toward reducing cost. Appendix~\ref{app:reasoning_coding} defines the categories and the rule by which a rationale is assigned to them.

Taken together, ReAct is the method to recommend when the objective matters more than reliably reaching feasibility, or when the inference cost of the language model is small relative to the simulation cost or is otherwise affordable.

\subsection{Larger models generally perform better, with the largest gains for ReAct}
\label{sec:res_scale}

Larger language models generally cost more to run, so here we quantify the trade-off between computational cost and performance across the 4B, 9B, and 27B sizes.

\begin{figure}[t]
    \centering
    \includegraphics[width=\linewidth]{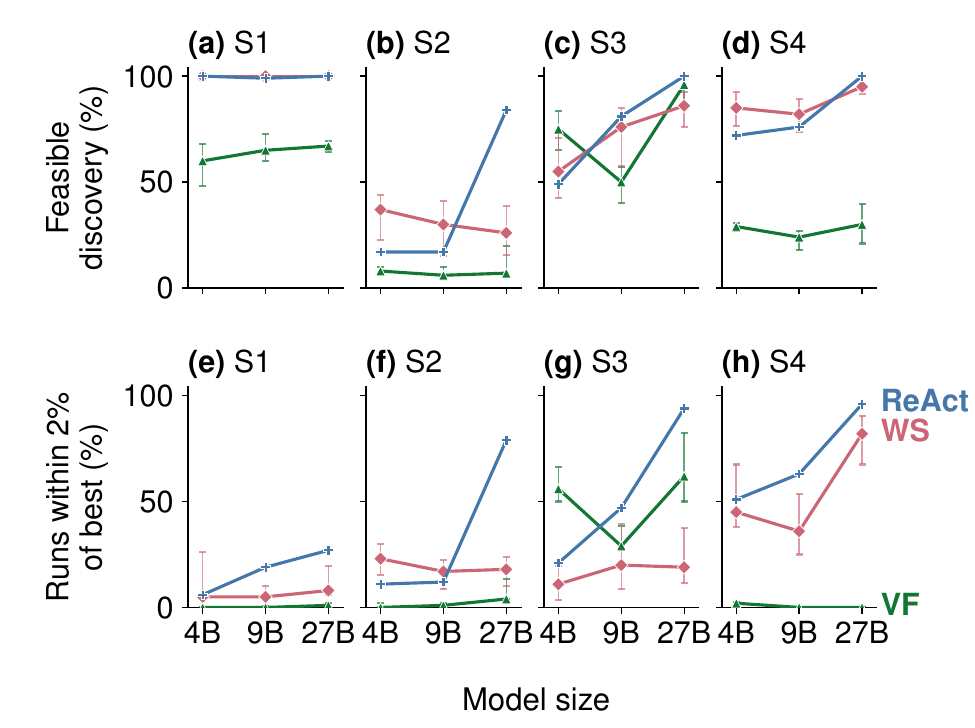}
    \caption{Feasible discovery (a)--(d) and near-best attainment (e)--(h) against model size. Points are medians across the configurations that share a model size and whiskers span their interquartile range.}
    \label{fig:R6}
\end{figure}

Figure~\ref{fig:R6} shows feasible discovery in panels (a) to (d) and the fraction of runs at a gap of 2\% or less in panels (e) to (h) against model size. Each point is the median across the configurations that share a model size, and the whiskers span their interquartile range. Both feasibility and optimality generally increase with model size, but the effect is by far the clearest for ReAct, and it is not always monotone for WS and for VF. We explain this by how much of the search each method gives to the language model. ReAct calls the language model at every step, so its dependence on model quality is the strongest. Therefore, when performance is the priority, ReAct with a large model can be the strongest choice. WS and VF use a downstream optimizer for the remaining search, so results depend on the interaction of the LLM and the numerical solver. The near-best attainment of VF in S3, for example, has a configuration median of 56\% at 4B, 29\% at 9B, and 62\% at 27B.

\begin{figure}[t]
    \centering
    \includegraphics[width=\linewidth]{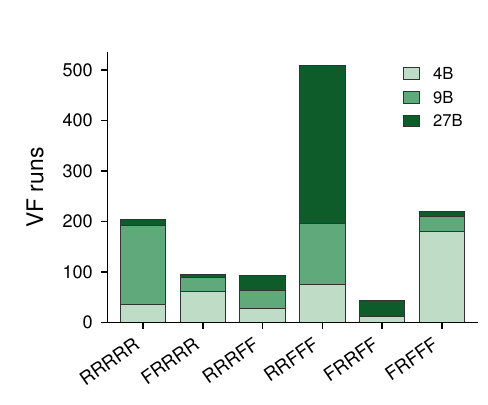}
    \caption{Variable-fixing masks selected in S3 by model size. Bars show the number of runs selecting each mask. R and F denote released and fixed variables in the order $(x^{\mathrm{C}},x^{\mathrm{T}},x^{\mathrm{S}},x^{\mathrm{M}},x^{\mathrm{R}})$.}
    \label{fig:R6b}
\end{figure}

The masks that each model size selects explain this non-monotone trend for VF. Figure~\ref{fig:R6b} shows the masks selected in S3 by model size. The 9B model released all five variables in 158 runs. The 4B model concentrated 181 runs on FRFFF, which releases only $x^{\mathrm{T}}$, and the 27B model concentrated 313 runs on RRFFF, which releases $x^{\mathrm{C}}$ and $x^{\mathrm{T}}$ together. The 4B model converts 168 of its 181 FRFFF runs into designs within 2\% of the best-known cost, which is a higher rate than the 238 of 313 that the 27B model achieves on RRFFF. However, the 4B model also places many runs on masks that never reach a near-best design. A more restrictive mask is therefore more efficient when it is correct, and the larger model does better overall because it selects a correct mask more consistently.

\begin{table}[t]
\caption{Median inference cost by strategy and model size}
\label{tab:compute}
\centering
\setlength{\tabcolsep}{4pt}
\begin{tabular}{lrrrrrr}
\toprule
& \multicolumn{3}{c}{Tokens/run} &
\multicolumn{3}{c}{Time/run (s)} \\
\cmidrule(lr){2-4}\cmidrule(lr){5-7}
Strategy & 4B & 9B & 27B & 4B & 9B & 27B \\
\midrule
VF    & 6,145   & 6,140   & 5,750   & 17  & 18  & 37 \\
WS    & 12,935  & 13,988  & 14,480  & 60  & 86  & 413 \\
ReAct & 775,090 & 459,364 & 254,678 & 474 & 438 & 592 \\
\bottomrule
\end{tabular}
\end{table}

Table~\ref{tab:compute} compares token use and wall-clock time. The 4B, 9B, and 27B models were served on one, two, and four GPUs, respectively, so the reported time reflects this allocation as much as the model itself and should be read as a relative indication. Appendix~\ref{app:compute} gives the full computational setup.

The number of tokens used for LLM inference shows the opposite trend. For ReAct the 27B model consumes fewer tokens per run than the smaller ones, with medians of 775,090 at 4B and 254,678 at 27B. If we calculate the number of tokens per call, it is similar across the model sizes. Thereby, different model sizes do not significantly change the number of tokens required to calculate, but since smaller models often propose candidates that repeat earlier ones, they become inefficient in terms of tokens spent.

All the results discussed in this section pool the two semantic-information levels, because supplying additional engineering information in the prompt (L2) produced no clear difference in the optimization outcomes or the redesign decisions. This suggests that the way such information is supplied to the model needs further study. Appendix~\ref{app:extended} gives the details.

\subsection{Implications}
\label{sec:discussion}

For practical redesign problems, the three WAFER strategies serve different functions. Each strategy applies the language model at a different point of the search, and the structure of the redesign problem determines which approach is most effective. Variable fixing changes the dimension of the search. Warm starting directs the search toward promising regions, and it performs consistently across all four scenarios. ReAct can change the search strategy at every step and is able to continue improving the objective after feasibility has been secured.

This suggests a simple selection rule. When the requirement change maps onto one or two design variables that can be identified with confidence, variable fixing gives the largest improvement at the lowest computational cost. When the promising region can be described but the responsible variables cannot be isolated, warm starting is the more reliable choice. When the objective still has room to improve after feasibility is reached, ReAct can be effective, and a larger ReAct model is preferable when the inference cost is not the main concern. Importantly, a larger iteration budget could change these observed trends, since a larger iteration budget can affect the ability of numerical solvers to identify feasible solutions and approach global optimality.

\section{Conclusions}

We compared three ways of integrating a large language model inside a simulator-driven redesign loop (warm starting, variable fixing, and ReAct search) against cold-start and reference-warm numerical optimization. The comparison used an electric vehicle battery-pack design problem evaluated with the BatPaC simulator under four redesign scenarios that change material cost, packaging space, charging requirement, and required energy. The key findings of this study are as follows.

\begin{enumerate}
\item Language model guidance reaches a feasible redesign in fewer simulator evaluations and ends at a lower cost than the numerical baselines under the same budget. The gain is largest in the scenarios where the reference design becomes infeasible and the feasible region is difficult to locate from function values alone.

\item The most effective LLM-assisted method differs across redesign scenarios. The outcome is decided by the way each method uses the reference design, so the structure of the requirement change determines which method to apply.

\item The benefit of a larger language model grows with the extent to which the model controls the search. It was largest for ReAct and noisy for warm starting and variable fixing, which use a numerical optimizer for the remaining budget.
\end{enumerate}

\balance
This study has several limitations. The first is the scope of the testbed. We used one simulator and one product family. Applying the same comparison to a different simulator and to a design problem whose variables are coupled in a different way would show whether the three methods show the same pattern. Constraint handling in the numerical optimizers may also vary by implementation, and a different choice could change how they compare with the LLM-assisted methods.

The second is the measure of search effort. We focus on a budget of simulator evaluations, which is the appropriate unit when a single simulation is far more expensive than the reasoning or search algorithm compute needed to determine which points to simulate. In our setting the median BatPaC evaluation took 6.5 seconds while a ReAct run required several hundred seconds of inference, so the two costs are not separated by the margin that this choice assumes. Repeating the comparison on a practical engineering simulator that takes hours per evaluation would show how the methods compare when the simulation cost dominates.

The third is the language model itself. We used only Qwen3.5 at three sizes, and frontier models may have stronger performance. Other model families including frontier models, or frameworks such as fine-tuning or retrieval-augmented generation, could change the behavior of the three methods.

Overall, our results suggest that LLMs have strong potential to provide effective assistance in simulation-based engineering redesign efforts by augmenting or even replacing numerical solvers. Our experiments begin to identify the types of redesign context and requirement shifts on which each LLM-based approach performs best. More research comparing LLM-assisted redesign across a range of engineering domains, LLM models, integration approaches, and numerical solvers could help more fully establish best practices for LLM use in redesign.

\section*{Code and Data Availability}

The code, the exact prompts used in the experiments, and the scripts that reproduce every table and figure are available at \coderepo. The result tables and the raw run records are available at \url{https://huggingface.co/datasets/Chani-Song/wafer-jmd-release-data}. BatPaC 5.0 is available from Argonne National Laboratory \cite{knehr2022batpac}.

\section*{Acknowledgment} %

The authors thank the members of the BOARD (Board of Agents for R\&D) research team for helpful discussions and feedback. This work was funded through the Collaborative Grand Challenge Initiative by the Department of Mechanical Engineering at Carnegie Mellon University.

\clearpage
\nobalance
\appendix
\section{Preliminary Reference-Design Search}
\label{app:source_grid}

The common reference design was selected using a deterministic discrete grid search under the reference requirements defined in Eq.~\eqref{eq:battery_source_problem}. The preliminary search grid is given in Table~\ref{tab:source_grid}.

\begin{table}[h]
\caption{Preliminary grid used to select the common reference design}
\label{tab:source_grid}
\centering
\setlength{\tabcolsep}{3pt}
\begin{tabular}{lll}
\toprule
Variable & Search range or values & Spacing \\
\midrule
$x^{\mathrm{C}}$
& LFP, NMC811
& categorical \\

$x^{\mathrm{T}}$ (NMC811)
& $\{100,120,140,150,155\}$~$\mu$m
& nonuniform \\

$x^{\mathrm{T}}$ (LFP)
& $\{100,120,140,150,160,165\}$~$\mu$m
& nonuniform \\

$x^{\mathrm{S}}$
& $14$--$30$
& $1$ \\

$x^{\mathrm{M}}$
& $2$--$4$
& $1$ \\

$x^{\mathrm{R}}$
& $2$--$6$
& $1$ \\
\bottomrule
\end{tabular}
\end{table}

\section{Prompts and Model Configuration}
\label{app:prompts}

Each prompt consists of a fixed system message and a structured user message. The user message contains the redesign problem definition, the reference design and its evaluated reference-condition result, the change from the reference to the redesign condition, the L1 or L2 semantic information, the strategy-specific instructions, and the required output format. The system message is identical under L1 and L2.

All three strategies use the same sampling settings. Requests are sent to a local vLLM endpoint with an OpenAI-compatible chat interface at temperature 0.7, top-$p$ 0.8, and a limit of 4096 generated tokens, with a context length of 32,768 tokens. The generation seed equals the run index. Structured output is enforced by the endpoint using a strict JSON schema. A response is rejected without consuming the simulator budget if it cannot be parsed, violates the schema, repeats an already evaluated candidate, or falls outside the variable bounds. In warm starting and variable fixing, a rejected response triggers one retry with the rejected output, the validation error, and the information needed to correct it, such as the chemistry-specific thickness bounds. A run whose retry also fails ends without a decision. In ReAct, a response that cannot be parsed or violates the schema triggers one repair request. A repeated or out-of-bounds candidate is instead returned to the agent as the observation for its next step. ReAct additionally allows at most 60 agent steps per run.

The complete system prompts, strategy instructions, optimizer-specific initialization instructions, ReAct action instructions, JSON output schemas, and the Level~1 and Level~2 semantic texts are reproduced verbatim in the code repository (\coderepo).

\section{L2 Engineering-Information Comprehension Check}
\label{app:l2_comprehension}

We conducted a separate comprehension test to verify that the engineering information supplied in the L2 prompt was interpretable by Qwen3.5-4B, the smallest model used in the main experiments. The test covered all four redesign scenarios with three independent responses per scenario, yielding 12 responses in total. The responses were generated locally with a 4-bit MLX version of Qwen3.5-4B at temperature 0.7 and top-$p$ 0.8.

The 16 questions asked the model to recover and interpret the engineering relationships supplied in L2 (Questions 1--11), to connect them to the five design variables in each redesign scenario (Question 12), to separate quantities that can be inferred from the prompt from those that require BatPaC evaluation (Questions 13--14), and to identify uncertain interpretations and separate supported conclusions from inference (Questions 15--16).

Across the 12 responses, Questions 1--11 were answered correctly in nearly all cases. One response contained a minor equation-transcription error, while the direction of the corresponding engineering relationship remained correct. For Questions 13--14, all responses correctly identified simulator-dependent quantities that could not be known before BatPaC evaluation. Responses to Question 12 generally connected the supplied relationships to the appropriate redesign variables, although several included additional design-level inferences beyond information explicitly stated in L2.

The L2 engineering-information text used in this test was identical to the Level~2 prompt. The four scenario-specific contexts, the 16 questions, and all 12 responses are reproduced in the code repository.

\begin{table}[t]
\caption{Feasible designs on the preliminary grid of Appendix~\ref{app:source_grid}. S1 is omitted because its reference design remains feasible.}
\label{tab:si_grid}
\centering
\setlength{\tabcolsep}{4pt}
\begin{tabular}{lrl}
\toprule
Sc. & Feasible & Shared by every feasible grid design \\
\midrule
S2 & 55  & $x^{\mathrm{R}}=3$ and $x^{\mathrm{M}}\geq3$ \\
S3 & 36  & $x^{\mathrm{T}}=100~\mu$m, the lowest thickness in the grid \\
S4 & 481 & $x^{\mathrm{M}}\geq3$ \\
\bottomrule
\end{tabular}
\end{table}

\section{Numerical Optimizer and Simulator Configuration}
\label{app:optimizer_settings}

All numerical search ran under Python 3.11.15 with NumPy 2.4.6 and SciPy 1.17.1. Bayesian optimization used Ax 1.3.1 with BoTorch 0.18.1, GPyTorch 1.15.2, and PyTorch 2.12.1. The genetic algorithm used pymoo 0.6.2. Mesh adaptive direct search used NOMAD 4.5.1 through the PyNomad 2.2 interface. Workbook editing used openpyxl 3.1.5.

Every optimizer receives the same penalized scalar objective. Let $G$ denote the sum of positive constraint violations. Each violation is normalized by the magnitude of its bound, except the fast-charge thickness limit of S3. That bound is computed by BatPaC for each candidate, and its violation is normalized by the 115~$\mu$m width of the widest thickness range. Feasible designs use the reported cost directly, whereas infeasible designs and simulator failures are assigned penalized losses:
\begin{equation}
\ell(\mathbf{x})=
\begin{cases}
C(\mathbf{x}), & \text{if feasible},\\
C(\mathbf{x})+\lambda_{\mathrm{base}}+\lambda_{\mathrm{gap}}G, & \text{if infeasible},\\
C_{0}+\lambda_{\mathrm{base}}+\lambda_{\mathrm{gap}}\max(G,1), & \text{if failed},
\end{cases}
\label{eq:penalty}
\end{equation}
where $\lambda_{\mathrm{base}}=10^{6}$ and $\lambda_{\mathrm{gap}}=10^{5}$. An evaluation fails when BatPaC raises an error or returns a nonfinite value, and $C_{0}$ is the cost it returns, or zero when none exists.

Cathode thickness is represented internally as a chemistry label together with a normalized coordinate $u_T\in[0,1]$, which maps to 50--165~$\mu$m for LFP and 50--155~$\mu$m for NMC811. Before each simulator call, a shared repair step clips thickness to the chemistry-specific bounds and converts the three topology variables with a rounding to the nearest integer followed by clipping to their bounds.

Chemistry is declared as an unordered categorical choice in Ax and pymoo and as an integer code in NOMAD. The topology variables are declared as integer ranges in all four methods.

\begin{table}[t]
\caption{Effect of the supplied engineering information. (a) Optimization outcomes under L1 and L2. (b) Redesign decisions under L1 and L2, pooled over scenarios.}
\label{tab:semantic}
\centering

\textbf{(a) Optimization outcomes}\par\vspace{2pt}
\setlength{\tabcolsep}{3pt}
\begin{tabular}{llrrrrrr}
\toprule
& &
\multicolumn{2}{c}{Discovery (\%)} &
\multicolumn{2}{c}{Gap $\leq2\%$ (\%)} &
\multicolumn{2}{c}{1st feas.} \\
\cmidrule(lr){3-4}\cmidrule(lr){5-6}\cmidrule(lr){7-8}
Strategy & Sc. & L1 & L2 & L1 & L2 & L1 & L2 \\
\midrule
VF & S1 & 61 & 63 & 0  & 0  & 2  & 2 \\
   & S2 & 11 & 12 & 6  & 5  & 13 & 13 \\
   & S3 & 72 & 71 & 54 & 51 & 3  & 3 \\
   & S4 & 28 & 27 & 2  & 1  & 10 & 10 \\
\addlinespace
WS & S1 & 99 & 99 & 10 & 12 & 1 & 1 \\
   & S2 & 30 & 29 & 20 & 18 & 10 & 7 \\
   & S3 & 67 & 71 & 19 & 20 & 6 & 3 \\
   & S4 & 87 & 83 & 64 & 47 & 2 & 6 \\
\addlinespace
ReAct & S1 & 100 & 99 & 23 & 12 & 3 & 3 \\
      & S2 & 37  & 41 & 31 & 37 & 8 & 8 \\
      & S3 & 71  & 83 & 47 & 61 & 3 & 2 \\
      & S4 & 83  & 82 & 65 & 75 & 2 & 3 \\
\bottomrule
\end{tabular}

\vspace{6pt}
\textbf{(b) Redesign decisions}\par\vspace{2pt}
\setlength{\tabcolsep}{6pt}
\begin{tabular}{lrr}
\toprule
Decision & L1 & L2 \\
\midrule
VF: release $x^{\mathrm{C}}$ (\%) & 72.2 & 66.8 \\
VF: release $x^{\mathrm{T}}$ (\%)             & 83.6 & 90.6 \\
VF: release $x^{\mathrm{S}}$ (\%)             & 60.5 & 58.5 \\
VF: release $x^{\mathrm{M}}$ (\%)  & 54.9 & 56.0 \\
VF: release $x^{\mathrm{R}}$ (\%)             & 54.2 & 56.2 \\
VF: required variable released (\%)  & 86.3 & 86.3 \\
\addlinespace
WS: reference-design candidate (\%)      & 13.0 & 11.2 \\
WS: variables changed                & 1.9  & 2.1 \\
WS: feasible candidate (\%)          & 24.8 & 21.4 \\
WS: distance from reference             & 0.1  & 0.2 \\
\bottomrule
\end{tabular}

\end{table}

For Bayesian optimization, the surrogate, kernel, acquisition function, and input transforms are the automatic selections of Ax 1.3.1. The genetic algorithm uses the mixed-variable implementation of pymoo with a population of five and terminates after 20 evaluations. Selection, crossover, mutation, and duplicate elimination use the defaults of pymoo 0.6.2 for mixed variables and were not overridden. Mesh adaptive direct search is called through PyNomad with a seed, and its own evaluation limit is set above the 20-evaluation budget. The mesh, direction, and search settings use the defaults of NOMAD 4.5.1. All supplied warm-start candidates are evaluated first and the candidate with the lowest penalized objective becomes the single starting point. Random search draws chemistry with equal probability, thickness from a uniform distribution over the chemistry-conditional interval, and each topology variable from a discrete uniform distribution. For run index $s$ in $0,\ldots,49$ the generation seed is $s$ and the optimizer seed is $2000+s$.

The objective and constraints are evaluated with the BatPaC 5.0 workbook dated 2022-02-09, accessed through a fixed cell interface on its Dashboard sheet. Each evaluation copies the workbook to a temporary directory, edits the input cells, runs a separate headless LibreOffice process to recalculate, and records the result. Across the 12,400 runs, the median evaluation took 6.54 seconds.

A run ends when the language model does not return an output that satisfies its schema within the allowed retry. This happened in 119 of the 1,600 variable-fixing runs at 4B and in 62 at 9B, and in none at 27B. Warm starting ended this way in 23 of its 4,800 runs and ReAct in none of its 1,200 runs. These runs stay in the denominator of every rate reported in the paper, so a run that produced no decision counts as a failure to discover a feasible design.

\begin{figure}[t]
    \centering
    \includegraphics[width=\linewidth]{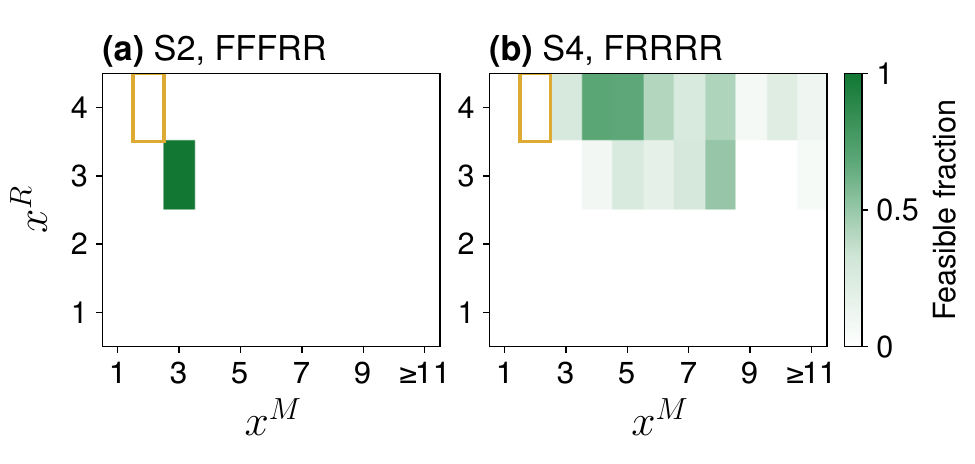}
    \caption{Feasible fraction of the accepted evaluations in each $(x^{\mathrm{R}},x^{\mathrm{M}})$ cell inside the VF subspace selected most often in S2 (FFFRR) and S4 (FRRRR) among the masks that fix at least one variable. The gold outline marks the reference design.}
    \label{fig:R3b}
\end{figure}

\section{Coding of the Reasoning Categories}
\label{app:reasoning_coding}

Every strategy returns a short rationale together with its structured action, and we evaluate the rationale using a deterministic rule over it. A category is recorded as present or absent for each rationale. Thus, every reported value is the fraction of rationales in which the category appears at least once. Table~\ref{tab:si_coding} writes each category as a logical rule over sets of terms.

\section{Extended Results}
\label{app:extended}

The preliminary grid search of Appendix~\ref{app:source_grid} found feasible designs in every redesign problem, as Table~\ref{tab:si_grid} shows.

Table~\ref{tab:semantic} compares the two semantic-information levels of Sec.~\ref{sec:semantic}. Performance does not differ noticeably. Some cells improve under L2 and others degrade, and the changes are small relative to the spread across configurations. No metric favors L2 consistently across the twelve combinations of strategy and scenario. The redesign decisions are also unchanged. Both the rate at which VF releases a required variable and the WS measures stay nearly identical under the two levels. The comprehension test of Appendix~\ref{app:l2_comprehension} shows that the smallest model reads the supplied relationships correctly. The reasoning categories of the rationales also change little between the two levels, as reported in the code repository. Qwen3.5 may already have battery-pack design knowledge at this level from pretraining, or supplying the information as one block of text may be an inefficient way for the model to use it, and this result is limited to the L2 text and delivery format used here.

The remaining figures and tables of this appendix provide the detail supporting the claims of Sec.~\ref{sec:results}. For the overall comparison, Table~\ref{tab:full_results} lists feasible discovery and near-best attainment for every configuration. The best reference-warm configuration reaches a higher feasible discovery than the worst configuration of every LLM-assisted method in all four problems, and the worst variable-fixing configuration falls below the best cold start in S1, S2, and S4. The aggregate advantage reported in the main text is therefore a statement about typical configurations, and does not make a claim about every configuration.

For the mechanism analysis, Table~\ref{tab:si_vf24}(a) and (b) repeat the mask analysis of Fig.~\ref{fig:R3a} for S2 and S4 and compare the downstream optimizers inside the VF subspaces selected most often. Figure~\ref{fig:R3b} shows the feasible fraction inside those subspaces, and Table~\ref{tab:exactsource}(c) compares warm-start runs with and without the reference design among the starting points. Table~\ref{tab:si_postfeas}(d) gives the post-feasibility conditional rates.

\section{Computational Setup}
\label{app:compute}

All experiments ran on a single Dell Precision 7960 Tower workstation under Ubuntu 22.04 LTS, with a 12-core Intel Xeon w5-3423 processor, 93\,GiB of system memory, and four NVIDIA RTX 4000 Ada Generation GPUs providing 20\,GB of memory each.

BatPaC evaluations ran on the CPU through a Python wrapper that drives the workbook with a headless LibreOffice 7.3.7.2 process. Evaluations within one optimization trajectory were executed sequentially.

\begin{table}[t]
\caption{Supplementary results for variable fixing, warm starting, and ReAct. (a) Feasible discovery and near-best attainment of variable fixing by mask in S2 and S4, in the format of Fig.~\ref{fig:R3a}. (b) Feasible discovery of variable fixing by downstream optimizer among the runs that selected FFFRR in S2 and FRRRR in S4, with 46 to 62 runs per optimizer. (c) Warm-start performance with and without the reference design among the starting points. (d) Conversion to a near-best design after the first feasible design in S2 to S4, for 27B ReAct and for the VF and WS configurations with the highest conversion among their feasible runs. $n_f$ is the number of runs that reached feasibility and Conv.\ is the percentage of those runs that also reached an optimality gap of 2\% or less. The next column gives, among the runs with at least ten accepted evaluations remaining, the percentage that reach an optimality gap of 2\% or less within ten evaluations after the first feasible design. $n_o$ is the number of runs that were feasible but not yet within 2\% at the first feasible design, and the last column is the percentage of those runs that reached 2\% later. The scripts in the code repository reproduce these rates for every configuration.}
\label{tab:si_vf24}
\label{tab:exactsource}
\label{tab:si_postfeas}
\centering

\textbf{(a) Variable fixing: outcome by mask}\par\vspace{1pt}
{\setlength{\tabcolsep}{3pt}
\begin{tabular}{lrrr@{\hskip 10pt}lrrr}
\toprule
\multicolumn{4}{c}{S2} & \multicolumn{4}{c}{S4} \\
\cmidrule(r{6pt}){1-4}\cmidrule(l){5-8}
Mask & $n$ & \shortstack{Disc.\\(\%)} & \shortstack{Gap $\leq2\%$\\(\%)} & Mask & $n$ & \shortstack{Disc.\\(\%)} & \shortstack{Gap $\leq2\%$\\(\%)} \\
\midrule
\texttt{RRRRR} & 359 & 9 & 1 & \texttt{RRRRR} & 536 & 36 & 0 \\
\texttt{FRRRR} & 123 & 11 & 0 & \texttt{FRRRR} & 193 & 41 & 0 \\
\texttt{FRFRR} & 109 & 15 & 8 & \texttt{RRRFF} & 175 & 0 & 0 \\
\texttt{FFRRR} & 67 & 33 & 7 & \texttt{FFRRR} & 73 & 53 & 10 \\
\texttt{RRRFF} & 67 & 0 & 0 & \texttt{RRFFF} & 72 & 0 & 0 \\
\texttt{FFFRR} & 234 & 21 & 21 & \texttt{FRRFF} & 22 & 0 & 0 \\
\texttt{RRFFF} & 79 & 0 & 0 & \texttt{FFRFF} & 45 & 0 & 0 \\
\texttt{FRFFF} & 60 & 0 & 0 &  & & & \\
\texttt{FFRFF} & 40 & 0 & 0 &  & & & \\
\bottomrule
\end{tabular}}

\vspace{3pt}
\textbf{(b) Variable fixing: discovery (\%) by optimizer}\par\vspace{1pt}
\begin{tabular}{llrrrr}
\toprule
Sc. & Mask & BO & MADS & GA & Rand. \\
\midrule
S2 & \texttt{FFFRR} & 4 & \textbf{69} & 7 & 6 \\
S4 & \texttt{FRRRR} & \textbf{54} & 43 & 29 & 39 \\
\bottomrule
\end{tabular}

\vspace{3pt}
\textbf{(c) Warm starting: with and without the reference design}\par\vspace{1pt}
\begin{tabular}{llrrrr}
\toprule
Sc. & Ref. &
$n$ &
\shortstack{Disc.\\(\%)} &
\shortstack{1st\\feas.} &
\shortstack{Gap $\leq2\%$\\(\%)} \\
\midrule
S1 & No  & 477   & 97.7  & 2 & 10.3 \\
   & Yes & 722   & 100.0 & 1 & \textbf{11.5} \\
\addlinespace
S2 & No  & 1,004 & 28.7 & 8 & 18.3 \\
   & Yes & 191   & 33.0 & 8 & \textbf{22.5} \\
\addlinespace
S3 & No  & 688 & 70.6 & 3 & \textbf{19.9} \\
   & Yes & 508 & 66.9 & 8 & 18.7 \\
\addlinespace
S4 & No  & 873 & 88.2 & 2 & \textbf{62.1} \\
   & Yes & 314 & 79.9 & 8 & 40.1 \\
\bottomrule
\end{tabular}

\vspace{3pt}
\textbf{(d) Conversion after the first feasible design}\par\vspace{1pt}
{\setlength{\tabcolsep}{3.5pt}
\begin{tabular}{llrrrrr}
\toprule
Sc. & Configuration & $n_f$ & \shortstack{Conv.\\(\%)} & \shortstack{Within 10\\(\%)} & $n_o$ & \shortstack{Conv.\\(\%)} \\
\midrule
S2 & ReAct 27B & 84  & 94 & 94 & 7   & 29 \\
   & VF 27B    & 72  & 79 & 53 & 17  & 12 \\
   & WS 4B     & 129 & 70 & 83 & 44  & 11 \\
\addlinespace
S3 & ReAct 27B & 100 & 94 & 93 & 53  & 89 \\
   & VF 4B     & 295 & 80 & 75 & 179 & 66 \\
   & WS 9B     & 285 & 33 & 30 & 248 & 23 \\
\addlinespace
S4 & ReAct 27B & 100 & 96 & 94 & 45  & 91 \\
   & VF 4B     & 108 & 7  & 8  & 104 & 4  \\
   & WS 27B    & 381 & 84 & 86 & 105 & 41 \\
\bottomrule
\end{tabular}}
\end{table}

\begin{table}[t]
\caption{Coding rules for the seven reasoning categories. Terms are matched without regard to case, sentences are split at a period, question mark, exclamation mark, or semicolon, and $\exists s$ means that one sentence of the rationale satisfies the condition. $a \rightarrow_{n} b$ means that a term of $b$ follows a term of $a$ within $n$ characters. The term sets are listed below the rules.}
\label{tab:si_coding}
\centering
\begin{tabular}{@{}p{0.22\columnwidth}p{0.72\columnwidth}@{}}
\toprule
Category & Rule \\
\midrule
Engineering mechanism & $\exists s:\ M(s) \wedge R(s)$ \\
Optimization history & $\exists s:\ H(s) \wedge N(s)$ \\
Constraint targeted & $K \vee (C \rightarrow_{60} A) \vee (E \rightarrow_{35} L) \vee T$ \\
Objective targeted & $(V \rightarrow_{40} O) \vee (\{\text{cost, objective}\} \rightarrow_{50} V') \vee (\text{best feasible} \rightarrow_{50} \text{cost})$ \\
Reference design & $S \vee (\text{restore} \rightarrow_{40} \text{source})$ \\
Exploration and uncertainty & $X$ \\
Generic & $G \wedge \neg(\text{any of the six categories above})$ \\
\midrule
Term set & Terms \\
\midrule
$M$ & thickness, electrode, cathode, transport, diffusion, impedance, fast charging, series, voltage, cell count, capacity, energy, kWh, modules per row, row(s), topology, pack width, footprint, geometry, chemistry, LFP, NMC, material cost, power, current, thermal, cell area, c\_max, e\_min, mpr, and the five variable names \\
$R$ & therefore, thus, hence, because, due to, lead(s) to, result(s) in, so that, in order to, to meet, satisfy, achieve, reduce, increase, maintain, or preserve, must, need to, require(s), will increase, decrease, reduce, raise, lower, improve, or preserve \\
$H$ & previous, prior, last, recent, earlier, so far, trajectory, iteration, previously evaluated \\
$N$ & a numeral, higher, lower, increase, decrease, improve, worsen, trend \\
$K$ & active, violated, binding, failed, or failing, followed by constraint or limit \\
$C$, $A$ & $C$: constraint, violation, limit. $A$: increase, decrease, reduce, raise, lower, change, keep, try, propose, must, need \\
$E$, $L$ & $E$: exceed(s), below, above. $L$: limit, minimum, maximum, constraint \\
$T$ & to meet, to satisfy, to restore \\
$V$, $O$ & $V$: reduce, lower, minimize, improve, with $V'$ adding better. $O$: cost, objective, \$/kWh, cost\_per\_kWh \\
$S$ & source followed by design, reference, point, configuration, topology, or architecture, revert, return to (the) source or baseline \\
$X$ & uncertain, unknown, hypothesis, explore, test whether, if, a, or the, try a, an, the, changing, increasing, decreasing, or reducing, alternative, different design, configuration, chemistry, or topology \\
$G$ & next step, next candidate, next design, optimizer, continue, adjust \\
\bottomrule
\end{tabular}
\end{table}

\begin{table*}[p]
\centering
\small
\caption{Feasible discovery and near-best attainment of every configuration. Each entry is the percentage of the 50 runs of a configuration, counting runs that ended without a usable decision as failures. Gap is attainment of an optimality gap of 2\% or less relative to the best-known cost of Table~\ref{tab:bestdesign}. Shaded cells are significantly better than the reference-warm baseline under the test of Table~\ref{tab:strategy}.}
\label{tab:full_results}
\begin{tabular}{llllrrrrrrrr}
\toprule
 & & & & \multicolumn{2}{c}{S1} & \multicolumn{2}{c}{S2} & \multicolumn{2}{c}{S3} & \multicolumn{2}{c}{S4} \\
\cmidrule(lr){5-6}\cmidrule(lr){7-8}\cmidrule(lr){9-10}\cmidrule(lr){11-12}
Method & Size & Level & Optimizer & Disc. & Gap & Disc. & Gap & Disc. & Gap & Disc. & Gap \\
\midrule
Cold &  &  & BO & 58 & 0 & 16 & 2 & 34 & 0 & 58 & 2 \\
 &  &  & GA & 26 & 2 & 6 & 0 & 10 & 0 & 24 & 0 \\
 &  &  & MADS & 28 & 0 & 8 & 2 & 18 & 2 & 32 & 2 \\
 &  &  & Random & 12 & 2 & 4 & 0 & 4 & 0 & 34 & 0 \\
\addlinespace
RW &  &  & BO & 100 & 0 & 26 & 10 & 44 & 0 & 68 & 14 \\
 &  &  & GA & 100 & 4 & 6 & 2 & 20 & 4 & 50 & 16 \\
 &  &  & MADS & 100 & 34 & 46 & 42 & 70 & 16 & 90 & 74 \\
 &  &  & Random & 100 & 2 & 4 & 0 & 4 & 0 & 34 & 0 \\
\addlinespace
VF & 4B & L1 & BO & 70 & 0 & 8 & 0 & \cellcolor[gray]{0.88}90 & \cellcolor[gray]{0.88}50 & 30 & 0 \\
 &  &  & GA & 68 & 0 & 6 & 0 & \cellcolor[gray]{0.88}68 & \cellcolor[gray]{0.88}58 & 12 & 0 \\
 &  &  & MADS & 64 & 0 & 6 & 0 & 82 & \cellcolor[gray]{0.88}76 & 30 & 6 \\
 &  &  & Random & 42 & 0 & 4 & 2 & \cellcolor[gray]{0.88}52 & \cellcolor[gray]{0.88}44 & 24 & 2 \\
 &  & L2 & BO & 68 & 0 & 10 & 0 & \cellcolor[gray]{0.88}88 & \cellcolor[gray]{0.88}64 & 28 & 0 \\
 &  &  & GA & 50 & 0 & 10 & 0 & \cellcolor[gray]{0.88}66 & \cellcolor[gray]{0.88}50 & 28 & 2 \\
 &  &  & MADS & 56 & 0 & 16 & 4 & 82 & \cellcolor[gray]{0.88}74 & 32 & 4 \\
 &  &  & Random & 42 & 0 & 8 & 2 & \cellcolor[gray]{0.88}62 & \cellcolor[gray]{0.88}54 & 32 & 2 \\
 & 9B & L1 & BO & 80 & 0 & 10 & 2 & 56 & \cellcolor[gray]{0.88}26 & 48 & 6 \\
 &  &  & GA & 62 & 2 & 6 & 0 & 38 & \cellcolor[gray]{0.88}24 & 18 & 0 \\
 &  &  & MADS & 62 & 0 & 12 & 6 & 52 & \cellcolor[gray]{0.88}40 & 24 & 0 \\
 &  &  & Random & 38 & 0 & 2 & 0 & \cellcolor[gray]{0.88}40 & \cellcolor[gray]{0.88}28 & 26 & 0 \\
 &  & L2 & BO & 80 & 0 & 10 & 2 & 70 & \cellcolor[gray]{0.88}38 & 30 & 0 \\
 &  &  & GA & 68 & 2 & 6 & 0 & \cellcolor[gray]{0.88}48 & \cellcolor[gray]{0.88}28 & 14 & 0 \\
 &  &  & MADS & 70 & 0 & 6 & 2 & 60 & \cellcolor[gray]{0.88}50 & 18 & 0 \\
 &  &  & Random & 54 & 0 & 2 & 0 & \cellcolor[gray]{0.88}40 & \cellcolor[gray]{0.88}30 & 24 & 0 \\
 & 27B & L1 & BO & 74 & 0 & 4 & 4 & \cellcolor[gray]{0.88}100 & \cellcolor[gray]{0.88}52 & 44 & 0 \\
 &  &  & GA & 66 & 2 & 6 & 4 & \cellcolor[gray]{0.88}92 & \cellcolor[gray]{0.88}72 & 12 & 0 \\
 &  &  & MADS & 68 & 0 & 56 & 52 & \cellcolor[gray]{0.88}100 & \cellcolor[gray]{0.88}100 & 38 & 4 \\
 &  &  & Random & 42 & 2 & 8 & 4 & \cellcolor[gray]{0.88}100 & \cellcolor[gray]{0.88}80 & 30 & 0 \\
 &  & L2 & BO & 80 & 0 & 0 & 0 & \cellcolor[gray]{0.88}96 & \cellcolor[gray]{0.88}38 & 48 & 0 \\
 &  &  & GA & 66 & 2 & 6 & 4 & \cellcolor[gray]{0.88}72 & \cellcolor[gray]{0.88}50 & 12 & 0 \\
 &  &  & MADS & 68 & 0 & 56 & 42 & \cellcolor[gray]{0.88}96 & \cellcolor[gray]{0.88}90 & 24 & 0 \\
 &  &  & Random & 58 & 2 & 8 & 4 & \cellcolor[gray]{0.88}68 & \cellcolor[gray]{0.88}50 & 30 & 0 \\
\addlinespace
WS & 4B & L1 & BO & 98 & 2 & 46 & 28 & \cellcolor[gray]{0.88}86 & 4 & 92 & \cellcolor[gray]{0.88}42 \\
 &  &  & GA & 98 & 8 & 26 & 22 & \cellcolor[gray]{0.88}48 & \cellcolor[gray]{0.88}30 & \cellcolor[gray]{0.88}88 & \cellcolor[gray]{0.88}66 \\
 &  &  & MADS & 100 & 26 & 44 & 36 & 50 & 14 & 96 & 78 \\
 &  &  & Random & 100 & 0 & 12 & 6 & 12 & 2 & \cellcolor[gray]{0.88}66 & \cellcolor[gray]{0.88}40 \\
 &  & L2 & BO & 100 & 0 & 38 & 18 & \cellcolor[gray]{0.88}86 & 8 & 82 & 32 \\
 &  &  & GA & 100 & 26 & \cellcolor[gray]{0.88}36 & \cellcolor[gray]{0.88}24 & \cellcolor[gray]{0.88}60 & 24 & \cellcolor[gray]{0.88}80 & \cellcolor[gray]{0.88}48 \\
 &  &  & MADS & 100 & 28 & 44 & 38 & 66 & 18 & 94 & 72 \\
 &  &  & Random & 94 & 0 & 12 & 8 & \cellcolor[gray]{0.88}26 & 0 & 64 & \cellcolor[gray]{0.88}22 \\
 & 9B & L1 & BO & 100 & 0 & 32 & 16 & \cellcolor[gray]{0.88}88 & 10 & 76 & 32 \\
 &  &  & GA & 100 & 6 & 28 & 20 & \cellcolor[gray]{0.88}72 & \cellcolor[gray]{0.88}44 & \cellcolor[gray]{0.88}82 & \cellcolor[gray]{0.88}52 \\
 &  &  & MADS & 100 & 22 & 48 & 30 & 80 & 38 & 92 & 64 \\
 &  &  & Random & 98 & 0 & 12 & 2 & \cellcolor[gray]{0.88}40 & 4 & \cellcolor[gray]{0.88}66 & \cellcolor[gray]{0.88}26 \\
 &  & L2 & BO & 100 & 6 & 40 & 18 & \cellcolor[gray]{0.88}96 & 12 & 88 & 22 \\
 &  &  & GA & 96 & 4 & 16 & 10 & \cellcolor[gray]{0.88}60 & \cellcolor[gray]{0.88}28 & \cellcolor[gray]{0.88}82 & 40 \\
 &  &  & MADS & 100 & 26 & 44 & 32 & 84 & \cellcolor[gray]{0.88}46 & 92 & 58 \\
 &  &  & Random & 98 & 0 & 12 & 4 & \cellcolor[gray]{0.88}50 & 4 & 40 & 4 \\
 & 27B & L1 & BO & 98 & 8 & 40 & 22 & \cellcolor[gray]{0.88}94 & \cellcolor[gray]{0.88}18 & \cellcolor[gray]{0.88}96 & \cellcolor[gray]{0.88}88 \\
 &  &  & GA & 100 & 8 & 16 & 12 & \cellcolor[gray]{0.88}86 & \cellcolor[gray]{0.88}42 & \cellcolor[gray]{0.88}100 & \cellcolor[gray]{0.88}98 \\
 &  &  & MADS & 100 & 34 & 36 & 30 & 86 & 20 & 100 & \cellcolor[gray]{0.88}98 \\
 &  &  & Random & 96 & 6 & 16 & 14 & \cellcolor[gray]{0.88}58 & 2 & \cellcolor[gray]{0.88}94 & \cellcolor[gray]{0.88}88 \\
 &  & L2 & BO & 100 & 6 & 38 & 22 & \cellcolor[gray]{0.88}92 & 14 & 92 & \cellcolor[gray]{0.88}60 \\
 &  &  & GA & 100 & 16 & 14 & 4 & \cellcolor[gray]{0.88}82 & \cellcolor[gray]{0.88}36 & \cellcolor[gray]{0.88}90 & \cellcolor[gray]{0.88}70 \\
 &  &  & MADS & 100 & 30 & 42 & 38 & \cellcolor[gray]{0.88}94 & 42 & 100 & 76 \\
 &  &  & Random & 100 & 2 & 10 & 0 & \cellcolor[gray]{0.88}56 & 4 & \cellcolor[gray]{0.88}90 & \cellcolor[gray]{0.88}60 \\
\addlinespace
ReAct & 4B & L1 & -- & 100 & 8 & 14 & 10 & 36 & 16 & 74 & 46 \\
 &  & L2 & -- & 100 & 4 & 20 & 12 & 62 & 26 & 70 & 56 \\
 & 9B & L1 & -- & 100 & 28 & 16 & 10 & 76 & 38 & 76 & 56 \\
 &  & L2 & -- & 98 & 10 & 18 & 14 & 86 & \cellcolor[gray]{0.88}56 & 76 & 70 \\
 & 27B & L1 & -- & 100 & 32 & \cellcolor[gray]{0.88}82 & \cellcolor[gray]{0.88}74 & \cellcolor[gray]{0.88}100 & \cellcolor[gray]{0.88}88 & 100 & \cellcolor[gray]{0.88}94 \\
 &  & L2 & -- & 100 & 22 & \cellcolor[gray]{0.88}86 & \cellcolor[gray]{0.88}84 & \cellcolor[gray]{0.88}100 & \cellcolor[gray]{0.88}100 & 100 & \cellcolor[gray]{0.88}98 \\
\bottomrule
\end{tabular}
\end{table*}

\selectlanguage{english}

\clearpage
\balance
\bibliographystyle{asmejour}   %

\bibliography{asmejour-sample} %

\end{document}